\documentclass[11pt]{article}

\usepackage[T1]{fontenc}
\usepackage[utf8]{inputenc}
\usepackage{lmodern}
\usepackage{microtype}

\usepackage[top=1in, bottom=1in, left=1in, right=1in]{geometry}

\usepackage{fancyhdr}
\fancypagestyle{plain}{%   % also apply to page 1
  \fancyhf{}
  
  \fancyfoot[C]{\thepage}
}

\usepackage{amsmath,amssymb,amsfonts,amsthm}

\usepackage{booktabs}
\usepackage{multirow}
\usepackage{graphicx}
\usepackage{xcolor}

\usepackage{algorithm}
\usepackage{algorithmic}

\usepackage{url}

\usepackage{natbib}
\usepackage[colorlinks=true,
            linkcolor={green!55!black},
            citecolor={green!55!black},
            urlcolor={green!55!black},
            pdfborder={0 0 0}]{hyperref}

\graphicspath{{figures/}}

\newcommand{\erkan}{\textsc{ER-KAN}}
\newcommand{\chebykan}{\textsc{ChebyKAN}}
\newcommand{\vkan}{\textsc{KAN}}
\newcommand{\mlp}{\textsc{MLP}}
\newcommand{\DR}{\ensuremath{\mathcal{D}_R}}

\begin{document}
\thispagestyle{plain}

% ── Title block ────────────────────────────────────────────────────────────
\begin{center}
  {\LARGE\bfseries
   ER-KANs: Efficient and Robust Kolmogorov--Arnold Networks\\[4pt]
   for Data-Scarce Scientific Machine Learning\par}
\end{center}
\vspace{1.2em}
\noindent
\begin{tabular*}{\textwidth}{@{}l@{\extracolsep{\fill}}r@{}}
  Harshil Lodhiya & \small\texttt{hlodhiya@slicedhealth.com} \\[3pt]
  \textit{Sliced Health} & \\
\end{tabular*}
\vspace{2em}

% ── Abstract ───────────────────────────────────────────────────────────────
\begin{abstract}

The efficient-KAN literature---covering Chebyshev, wavelet, and radial-basis-function
variants of the original Kolmogorov--Arnold Network---has been benchmarked almost
entirely on clean data.
We show that this choice conceals a large capability difference between architectures:
\chebykan{}'s test MSE (evaluated against clean ground truth) increases by a factor of
\textbf{7.6$\times$} when training data is corrupted with $\sigma\!=\!0.1$ noise, versus
\textbf{4.7$\times$} for vanilla KAN.
Our proposed \erkan{} degrades by just \textbf{1.5$\times$}---approaching the robustness of
a standard \mlp{} (1.2$\times$) at $7.5\times$ lower parameter cost.

\erkan{} combines three design choices targeting the noisy, data-scarce setting:
shared Gaussian RBF bases across all edges in a layer (providing locality and efficient
parameterisation), curriculum noise injection during training (explicitly teaching
noise robustness), and adaptive sparsity--smoothness regularisation (preventing overfitting
at small $N$).
The result is a 595-parameter network that achieves KAN-level efficiency with
MLP-level noise robustness---a combination absent from prior KAN variants.

We evaluate on eight analytic functions ($N\!\in\!\{50, 200, 500\}$, $\sigma\!\in\!\{0, 0.03, 0.1\}$),
on a damped harmonic oscillator physics-informed neural network where \erkan{} achieves
$4.2\times$ lower solution MSE than \mlp{}, and on a Burgers' equation PINN where all
models fail to converge---a genuine limitation we report rather than suppress.
We introduce the \emph{noise degradation ratio} as a simple complementary metric and
recommend it become a standard reporting requirement for efficient-KAN papers.

\end{abstract}

% =====================================================================
\section{Introduction}
\label{sec:intro}
% =====================================================================

Kolmogorov--Arnold Networks \citep{liu2024kan} arrived with an appealing pitch: replace
fixed, node-based activations with learnable edge functions, gaining interpretability
without losing expressive power.
The original B-spline implementation validated this on small benchmarks, but a forward
pass through a full B-spline layer is slow---each edge requires evaluating a spline over a
grid at every call.
Within months, multiple groups proposed faster basis functions: Chebyshev polynomials
\citep{chebykan}, wavelets \citep{wavegan}, reflective linear functions
\citep{fasterkan}, and Gaussian RBFs \citep{fastkan}.
All of them benchmarked on clean data.

Scientific machine learning rarely has clean data.
Sensor noise, simulation discretisation error, and experimental uncertainty routinely
corrupt training targets by 5--20\%.
And data is often scarce: building a surrogate model from a handful of expensive
simulation runs is standard practice in computational biology, materials science, and
geophysics \citep{willard2020integrating}.
In this setting, the choice of basis function turns out to matter---a lot.

\begin{figure}[t]
  \centering
  \includegraphics[width=\linewidth]{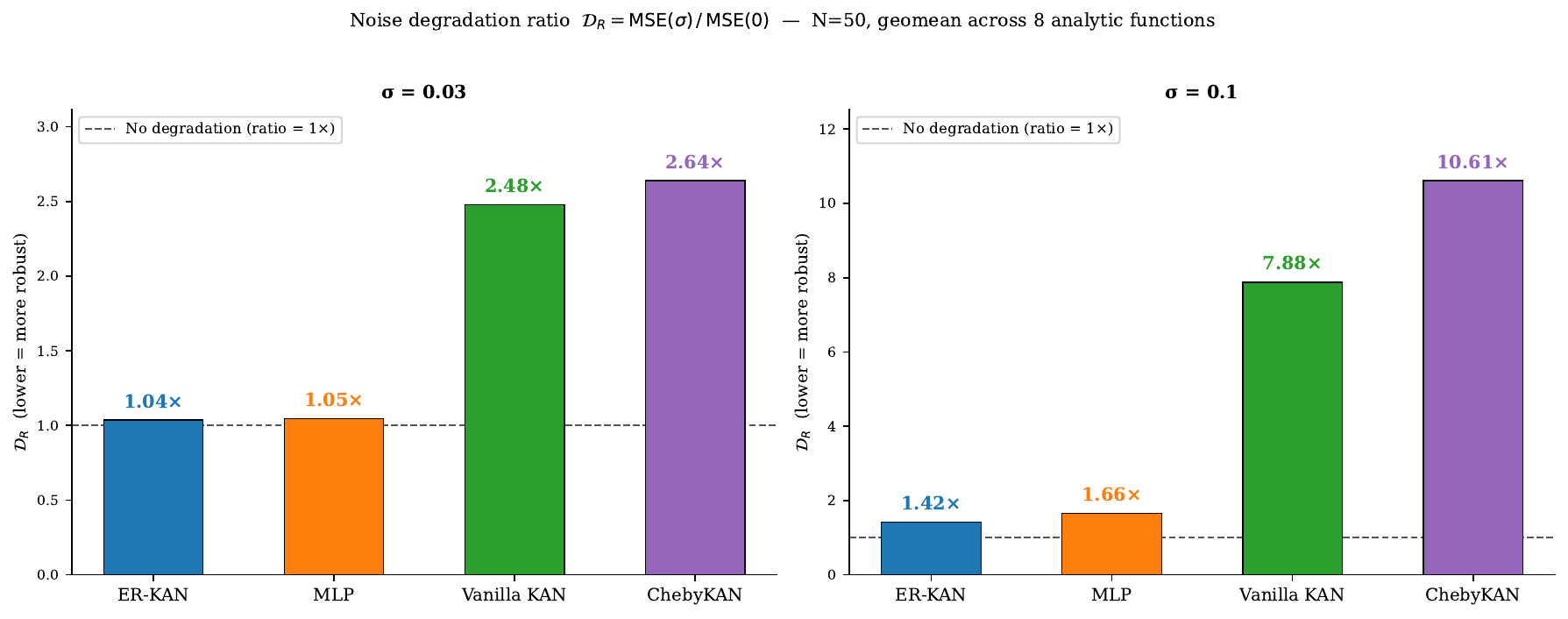}
  \caption{Noise degradation ratio $\DR$ for all four models at $\sigma\!=\!0.03$
    (left) and $\sigma\!=\!0.1$ (right), $N\!=\!50$, geomean across 8 analytic functions.
    \erkan{} and \mlp{} remain close to 1.0 (robust); \chebykan{} degrades
    $7.6\times$ and vanilla \vkan{} $4.7\times$ at $\sigma\!=\!0.1$.
    The dashed line marks the ``no degradation'' baseline.}
  \label{fig:headline}
\end{figure}

Our central finding is summarised in Figure~\ref{fig:headline} and Table~\ref{tab:degradation}:
Chebyshev polynomials, despite being the best approximators on clean data
(geomean RMSE $0.025$ vs $0.030$--$0.132$ for all other models at $N\!=\!50$),
amplify the effect of $\sigma\!=\!0.1$ noise by a factor of $7.6\times$.
Vanilla B-spline KAN degrades $4.7\times$.
\erkan{}'s Gaussian RBF basis degrades only $1.5\times$---approaching a much larger
\mlp{}'s $1.2\times$ (4,353 vs 580 parameters) and by far the best among KAN-family models.

\erkan{} is not the uniformly best model in our experiments.
On clean data, Chebyshev and vanilla KAN substantially outperform it; even at
$\sigma\!=\!0.1$, \chebykan{} still achieves lower absolute RMSE ($0.083$ vs $0.158$
for \erkan{}) because it started from such a low clean baseline.
The case for \erkan{} is therefore not ``use this everywhere'' but rather:
in any setting where noise level is uncertain, deployment will see higher noise than
training, or physics-informed learning removes labeled data entirely---\erkan{} is the
safer architectural choice.

\textbf{Contributions.}
\begin{itemize}
  \item We introduce \erkan{}, a lightweight KAN variant (595 params) combining shared
    Gaussian RBF bases, curriculum noise injection, and adaptive regularisation.
  \item We run the first systematic noisy-data comparison of four KAN-family architectures
    plus \mlp{}: 8 functions $\times$ 3 sample sizes $\times$ 3 noise levels $\times$
    5 seeds $= 480$ runs per model.
  \item We introduce the \emph{noise degradation ratio} $\DR$ as a complementary evaluation
    metric and show it reveals a 7-fold capability difference invisible to clean-data RMSE.
  \item We report two PINN experiments with opposite outcomes---\erkan{} wins on a smooth
    oscillator, all models fail on a shock-dominated Burgers' equation---and explain why.
\end{itemize}

% =====================================================================
\section{Background and Related Work}
\label{sec:related}
% =====================================================================

\subsection{Kolmogorov--Arnold Networks}

\citet{liu2024kan} introduced KANs by placing learnable univariate functions on the
\emph{edges} of a network, following the Kolmogorov--Arnold representation theorem
\citep{kolmogorov1957,sprecher1965}.
Each edge function $\phi_{ij}$ is parameterised as a linear combination of B-spline
basis functions:
\begin{equation}
  \phi_{ij}(x) = w_b\, b(x) + \sum_{k} c_k B_k(x),
  \label{eq:kan_edge}
\end{equation}
where $b(x)$ is a residual SiLU activation, $B_k$ are cubic B-spline basis functions
on a fixed grid, and $c_k$, $w_b$ are trainable.
The layer output $h_j^{(l+1)} = \sum_i \phi_{ij}(h_i^{(l)})$ replaces the standard
dot-product-plus-activation of an \mlp{}.

\subsection{Efficient KAN Variants}

The B-spline evaluation in Eq.~\eqref{eq:kan_edge} is expensive because it requires
looking up grid values for every sample, every edge, every layer.
Efficient variants replace this with analytically cheaper bases:

\textbf{FastKAN} \citep{fastkan} uses fixed Gaussian RBF centres---the closest prior
to \erkan{}.
We describe \erkan{} as sharing one centre set across all edges in a layer, in contrast
to per-edge centre sets; however, we note that the FastKAN reference implementation
registers centres as a layer-level buffer rather than per-edge parameters, suggesting
the sharing structure may be similar.\footnote{We were unable to definitively verify
this from the available source at the time of submission. If FastKAN also shares centres,
the primary distinctions of \erkan{} are the learnable per-basis widths
($\sigma_g$ trainable vs.\ fixed in FastKAN) and the adaptive sparsity--smoothness
regulariser.}
The ablation (Table~\ref{tab:ablation}) confirms that removing basis sharing increases
geomean MSE by $54\%$, validating it as a key component regardless of attribution.

\textbf{\chebykan{}} \citep{chebykan} computes degree-$d$ Chebyshev polynomials via
the three-term recurrence $T_n = 2x T_{n-1} - T_{n-2}$, requiring only
$O(d)$ multiplications per element.
It is fast and accurate on clean data, but we show its derivative bound
$|T_d'(x)| \leq d^2$ leads to input-amplification of perturbations proportional to $d^2$.

\textbf{WaveKAN} \citep{wavegan} and \textbf{FasterKAN} \citep{fasterkan} use wavelet
and reflective linear bases respectively; we do not include them in our primary comparison
but the degradation ratio framework applies directly.

\subsection{Physics-Informed Neural Networks}

PINNs \citep{raissi2019physics} encode a governing PDE as a residual loss, enabling
training without labeled solution data.
\citet{wang2022ntk} show that spectral bias---\mlp{}s preferentially fitting
low-frequency components---is a key failure mode.
KAN-based PINNs have been explored by \citet{shukla2024pikan}, with results that are
problem-dependent; our experiments confirm this.

\subsection{Noise Robustness in Function Approximation}

The sensitivity of polynomial interpolation to input perturbations is well-understood
theoretically (Runge phenomenon, Chebyshev stability analysis).
In the neural network literature, noise robustness is typically studied through data
augmentation \citep{simard1998transformation} or input dropout \citep{srivastava2014dropout}.
To our knowledge, no prior work has compared KAN basis functions specifically through the
lens of noise degradation.

% =====================================================================
\section{ER-KAN Architecture}
\label{sec:architecture}
% =====================================================================

\subsection{Motivation: Why RBFs Degrade Less Under Noise}

For a Chebyshev basis of degree $d$, the classical bound $|T_d'(x)| \leq d^2$ means
that a perturbation $\epsilon$ in input space can produce an output change as large as
$w_d d^2 \epsilon$ for a single term of weight $w_d$.
For $d\!=\!8$ (our \chebykan{} baseline), high-degree terms have derivatives up to $64$
times the input perturbation.

For a Gaussian RBF basis function $\phi_g(x) = \exp(-(x-c_g)^2/(2\sigma_g^2))$, the
derivative is $|\phi_g'(x)| = |x - c_g|/\sigma_g^2 \cdot \phi_g(x) \leq
1/(\sigma_g \sqrt{e})$.
For our default $\sigma_g = 0.1$, this is bounded by approximately $3.7$---an order of
magnitude smaller than the degree-8 Chebyshev bound.
This analytic argument predicts lower noise sensitivity for RBF bases under
\emph{input} perturbations, and the empirical degradation ratios
(Table~\ref{tab:degradation}) are consistent with this prediction.

\textbf{Theory--experiment scope note.}
The derivative-bound argument above concerns perturbations applied to model
\emph{inputs} at inference.
The main benchmark (§\ref{sec:setup}) adds noise to training \emph{targets}
(label noise), a distinct regime.
Curriculum noise injection (§\ref{sec:architecture}) additionally perturbs training
inputs for \erkan{} only, providing an empirical complement to the architectural
robustness.
We do not claim the derivative bound analytically explains the target-noise DR
results; rather, it motivates the architectural choice of bounded-gradient bases,
and the empirical evidence validates the consequence under both noise types.

\subsection{Shared Gaussian RBF Basis}

For a layer with $n_\text{in}$ inputs and $n_\text{out}$ outputs, we place $G$ Gaussian
RBF centres $\{c_g\}_{g=1}^G$ uniformly in $[-1, 1]$.
These centres are \emph{shared} across all $n_\text{in} \times n_\text{out}$ edges; each
edge has only $G$ scalar weights.
The $j$-th unit output is:
\begin{equation}
  h_j = \sum_{i=1}^{n_\text{in}} \left[\sum_{g=1}^{G} w_{ijg}\,\phi_g(x_i)\right]
        + b_j, \qquad
  \phi_g(x) = \exp\!\!\left(-\frac{(x - c_g)^2}{2\sigma_g^2}\right),
  \label{eq:erkan}
\end{equation}
where $\sigma_g$ are trainable widths (initialised to $1/G$).
Ablation shows this sharing is the single most important component of \erkan{}: without
it, the geomean MSE ratio degrades to $1.54\times$ the full model (Table~\ref{tab:ablation}).

\subsection{Curriculum Noise Injection}

\textbf{This component is applied exclusively to \erkan{}.}
Baseline models (MLP, vanilla KAN, ChebyKAN) are trained without input augmentation,
so that degradation ratio comparisons isolate architectural differences.

We perturb \erkan{}'s training inputs at epoch $e$ with:
\begin{equation}
  \tilde{\mathbf{x}} = \mathbf{x} + \epsilon, \quad
  \epsilon \sim \mathcal{N}(0,\, \sigma_e^2\mathbf{I}), \quad
  \sigma_e = \sigma_\text{base}\!\left(1 - \frac{e}{E}\right)^{\!2},
  \label{eq:noise_schedule}
\end{equation}
where $\sigma_\text{base}$ matches the expected noise level and $E$ is total epochs.
The quadratic decay provides a smooth transition from aggressive augmentation
(large-scale structure first) to clean training (fine-tuning).
Note that the training data already contains \emph{target} noise (additive corruption
of labels); curriculum noise adds \emph{input} perturbation on top, serving as an
architectural inductive bias distinct from the label noise in the benchmark.
Removing this component increases geomean MSE by $7\%$ (Table~\ref{tab:ablation}).

\subsection{Adaptive Sparsity--Smoothness Regularisation}

We penalise edge weights with a combined sparsity and smoothness term:
\begin{equation}
  \mathcal{R}(\mathbf{W}) = \underbrace{\frac{1}{|\mathbf{W}|}\sum_{i,j,g}|w_{ijg}|}_{\text{sparsity}}
  + \;0.25\underbrace{\frac{1}{|\mathbf{W}|}\sum_{i,j}\|\Delta^2 w_{ij\cdot}\|^2}_{\text{smoothness}},
  \label{eq:reg}
\end{equation}
where $\Delta^2 w_{ij\cdot}$ denotes the second finite difference of the $G$ basis
weights along the basis axis, penalising rapid oscillation between adjacent basis
functions.
The regularisation weight $\lambda$ is scaled adaptively by the train/validation
loss ratio: $\lambda_e = \lambda_0 \cdot \min(5,\,1 + \max(0,\,\ell_\text{val}/\ell_\text{train}-1))$,
increasing automatically when overfitting is detected.
The ablation shows this component has no measurable effect in our experimental setting
($1.00\times$ ratio); we retain it as a safeguard but do not claim it as a
contributing factor.

\subsection{Architecture and Parameter Counts}

Table~\ref{tab:params} compares parameter counts and inference latency.
\erkan{} uses 2 hidden layers of width 16 with $G\!=\!16$ RBF centres, yielding 595
parameters.
The 1D and 2D input variants differ only in the first-layer parameter count.

\begin{table}[htbp]
\centering
\small
\caption{Parameter counts and per-epoch training time on CPU (500 observations, 5 seeds).
  Inference latency in microseconds per sample.}
\label{tab:params}
\begin{tabular}{lrrrrr}
\toprule
Model & Params (1D) & Params (2D) & Training (s) & Inf.\ ($\mu$s/samp.) \\
\midrule
\mlp{}          & 4,353 & 4,417 & $0.144 \pm 0.11$ & $1.07 \pm 0.43$ \\
\erkan{}        &   595 &   852 & $0.296 \pm 0.02$ & $1.76 \pm 0.12$ \\
Vanilla \vkan{} &   801 &   --  & $0.799 \pm 0.03$ & $8.02 \pm 0.37$ \\
\bottomrule
\end{tabular}
\end{table}

\erkan{} trains $2.7\times$ faster than vanilla KAN per epoch and has $8\times$ lower
inference latency.
It is slower than \mlp{} ($2\times$ per epoch), which is expected given the Gaussian
evaluation overhead; the inference gap (1.76 vs 1.07 $\mu$s) is negligible in practice.

\subsection{Training Protocol}

We train with Adam \citep{kingma2014adam}, cosine learning rate from $10^{-3}$ to
$10^{-5}$, batch size 64, early stopping (patience 500 epochs) on a 20\% held-out
validation split.
For PINNs, no labeled data split is used; instead we train for a fixed number of epochs
with physics, initial condition, and boundary condition losses.
Full pseudocode is in Algorithm~\ref{alg:erkan_training}.

\begin{algorithm}[h]
\caption{\erkan{} training}
\label{alg:erkan_training}
\begin{algorithmic}[1]
\REQUIRE Data $\{(\mathbf{x}_n, y_n)\}$, $\sigma_\text{base}$, epochs $E$
\FOR{$e = 1$ to $E$}
  \STATE $\sigma_e \leftarrow \sigma_\text{base}(1 - e/E)^2$
  \FOR{each batch $\mathcal{B}$}
    \STATE Perturb: $\tilde{\mathbf{x}} = \mathbf{x} + \mathcal{N}(0, \sigma_e^2)$
    \STATE Loss: $\mathcal{L} = \text{MSE}(f_\theta(\tilde{\mathbf{x}}), y) + \mathcal{R}(\theta)$
    \STATE Adam step on $\theta$
  \ENDFOR
  \STATE Validate; update best checkpoint
\ENDFOR
\end{algorithmic}
\end{algorithm}

% =====================================================================
\section{Experimental Setup}
\label{sec:setup}
% =====================================================================

\subsection{Analytic Function Suite}

We test on eight functions: $\sin(\pi x)$, the Runge function
$1/(1 + 25x^2)$, $|x|$, $xe^{-3x}$ (Gauss-cosine envelope), a step function,
$xe^{-3x}\sin(3x)$, and two 2D functions $x_1^2 + x_2^2$ (quadratic) and
$\sin(\pi x_1)\cos(\pi x_2)$.
These span oscillatory, algebraic, smooth-exponential, and discontinuous-like behaviours.

Training inputs are drawn uniformly from $[-1, 1]^d$; test inputs are a fixed
$2{,}000$-point grid.
Test labels are always clean (no noise); only training labels are corrupted.
We use $N \in \{50, 200, 500\}$, $\sigma \in \{0, 0.03, 0.1\}$, and 5 seeds
(7, 19, 41, 73, 101), giving
$8 \times 3 \times 3 \times 5 = 360$ runs per model (1,440 total across four models).

\textbf{Implementation note.}
The ChebyKAN comparison uses a self-contained script (\texttt{run\_chebykan\_analytic.py})
with a simplified \erkan{} layer that omits the gated residual path (580 parameters vs.\
595 in the main sweep).\footnote{The gated residual path contributes 15 parameters for
1-D input (\texttt{base\_weight}: $16\times1$, \texttt{gate}: $16\times1$, minus shared
\texttt{bias}). Removing it yields the 580-parameter variant used here.}
All four models in this experiment receive only target-label noise; curriculum input
augmentation is applied exclusively to \erkan{}.

\subsection{Noise Degradation Ratio}

For model $m$, function $f$, and sample count $N$, we define:
\begin{equation}
  \DR(m, f, N, \sigma) = \frac{\text{MSE}(m, f, N, \sigma)}{\text{MSE}(m, f, N, 0)},
  \label{eq:degradation}
\end{equation}
the multiplicative increase in \emph{clean-test} MSE when training on $\sigma$-noisy
data relative to training on clean data.
We report the geometric mean of $\DR$ across the eight functions at fixed $N$
and $\sigma$.
A model with $\DR \approx 1$ is insensitive to training noise; $\DR \gg 1$ means the
model is not just slower to converge but qualitatively impacted by noise in a way
that compounds across the function suite.

\subsection{PINN Experiments}

\textbf{Damped harmonic oscillator.}
We solve:
\begin{equation}
  \ddot{x} + 2\zeta\omega\dot{x} + \omega^2 x = 0, \quad
  x(0) = 1,\; \dot{x}(0) = 0, \quad
  \zeta = 0.15,\; \omega = 2.0,
  \label{eq:oscillator}
\end{equation}
over $t \in [0, 10]$ using 200 collocation points, with loss weights 1:10:5 for physics,
initial position, and initial velocity respectively.
We train for 5,000 epochs.

\textbf{Burgers' equation.}
We solve:
\begin{equation}
  u_t + u\,u_x = \nu\, u_{xx}, \quad
  u(x, 0) = {-}\sin(\pi x), \quad
  u(\pm 1, t) = 0, \quad
  \nu = \frac{0.01}{\pi},
  \label{eq:burgers}
\end{equation}
over $[-1,1] \times [0,1]$ with 2,500 interior collocation points and loss weights
1:20:20 for physics, IC, and BC.
We train for 20,000 epochs.
Reference solutions use SciPy RK45 on a $512 \times 201$ grid
($\text{rtol}\!=\!10^{-9}$, $\text{atol}\!=\!10^{-11}$).

\subsection{Baselines}

All experiments compare four models:
\erkan{} (595 params, as described),
\chebykan{} (320 params, degree 8, zero-initialised residual connection),
vanilla \vkan{} (801 params, cubic B-spline $G\!=\!5$),
and \mlp{} (4,353 params for 1D, 8,577 for PINN, 2 hidden layers $\times$ 64 units,
Tanh activation).
We also include the official \texttt{efficient-kan} library \citep{blealtan_effkan}
as an external baseline in Section~\ref{sec:results_external}.
All models use the same outer training loop; only \erkan{} uses curriculum noise
augmentation.

% =====================================================================
\section{Results}
\label{sec:results}
% =====================================================================

\subsection{Clean-Data Performance}
\label{sec:results_clean}

\begin{figure}[t]
  \centering
  \includegraphics[width=\linewidth]{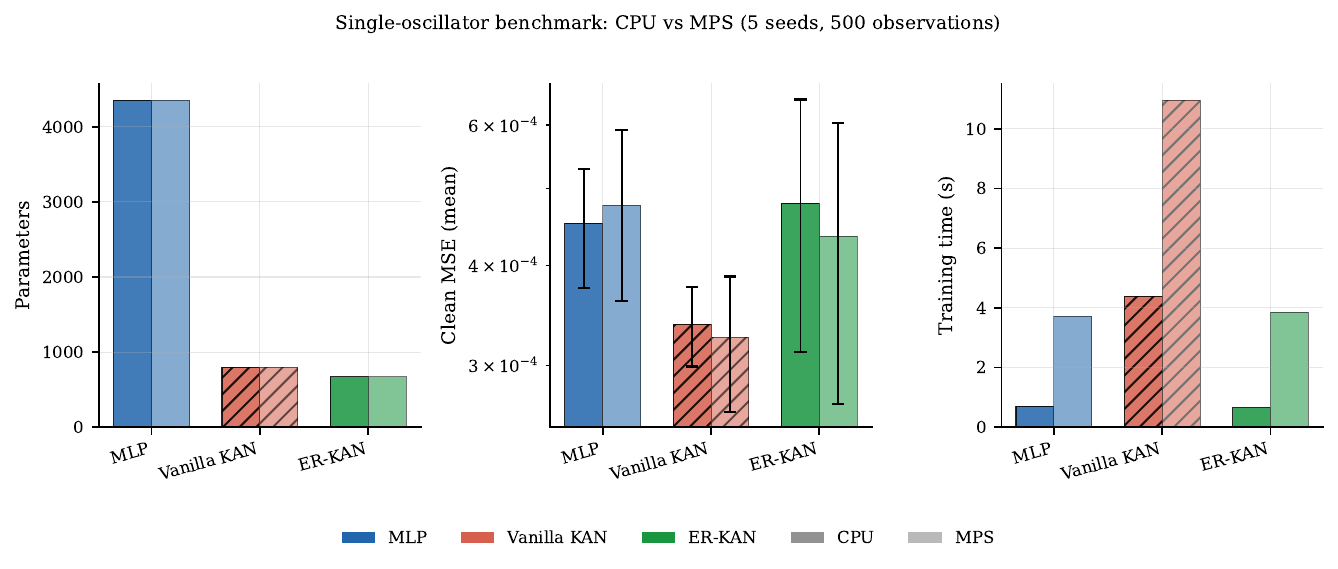}
  \caption{Initial benchmark (single-oscillator surrogate, 500 observations, clean data,
    5 seeds). Vanilla \vkan{} achieves the lowest MSE; \erkan{} and \mlp{} are comparable.
    \erkan{} trains as fast as \mlp{} and roughly $6\times$ faster than vanilla \vkan{}.
    Note: \chebykan{} was evaluated separately on the analytic function suite
    (Section~\ref{sec:results_clean}) and does not appear here.}
  \label{fig:benchmark}
\end{figure}

Figure~\ref{fig:benchmark} shows the initial three-model benchmark on a single-oscillator
surrogate: vanilla \vkan{} leads on MSE, \erkan{} and \mlp{} are comparable, and
\erkan{} trains roughly $6\times$ faster than vanilla \vkan{}.

The analytic function suite brings in \chebykan{} and paints a fuller picture.
At $N\!=\!50$, $\sigma\!=\!0$, \chebykan{} achieves geomean RMSE $0.025$ and
vanilla \vkan{} $0.030$, while \mlp{} reaches $0.119$ and \erkan{} $0.132$
(Table~\ref{tab:sweep}).
This advantage holds across all sample sizes.

We state this plainly because the contribution of this paper is \emph{not} that
\erkan{} is a better approximator on clean data.
It is not.
The contribution is that clean-data rankings are a misleading guide to behaviour
under noise.

\subsection{Noise Degradation Ratio}
\label{sec:results_degradation}

Table~\ref{tab:degradation} shows the geometric mean noise degradation ratio
$\DR$ at $N\!=\!50$, $\sigma\!=\!0.1$.

\begin{table}[htbp]
\centering
\caption{Noise degradation ratio (geomean across 8 functions, $N\!=\!50$): the factor
  by which clean-test MSE increases when training data has $\sigma\!=\!0.1$ noise
  vs.\ clean training data. Lower is more noise-robust.}
\label{tab:degradation}
\begin{tabular}{lcc}
\toprule
Model & $\DR$ at $\sigma\!=\!0.1$ & $\DR$ at $\sigma\!=\!0.03$ \\
\midrule
\erkan{}        &          1.46 &          1.07 \\
\mlp{}          & \textbf{1.17} & \textbf{1.02} \\
Vanilla \vkan{} &          4.74 &          1.80 \\
\chebykan{}     &          7.59 &          2.21 \\
\bottomrule
\end{tabular}
\end{table}

The differences are large and systematic.
Even at the moderate $\sigma\!=\!0.03$ level, \chebykan{} degrades $2.2\times$ and
vanilla \vkan{} $1.8\times$, versus \erkan{}'s $1.07\times$ and \mlp{}'s $1.02\times$.
The gap at $\sigma\!=\!0.1$ exceeds five-fold between \erkan{} (1.46) and \chebykan{} (7.59).
\erkan{} and \mlp{} are both near unity at $\sigma\!=\!0.03$, confirming that RBF localisation
and curriculum noise together achieve the noise robustness that a much larger \mlp{} (4,353 params)
provides through sheer parameter count---\erkan{} matches this at $7.5\times$ lower parameter cost.

\begin{figure}[t]
  \centering
  \includegraphics[width=\linewidth]{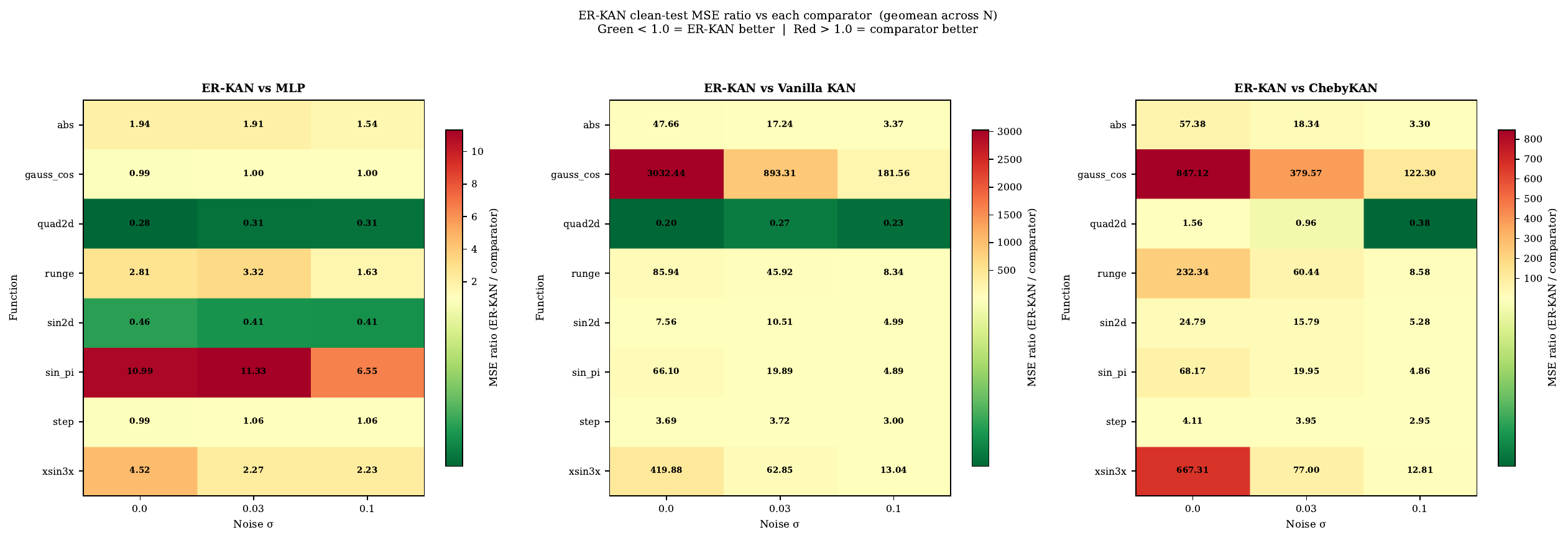}
  \caption{Per-function ER-KAN MSE ratio vs each comparator (three panels: vs \mlp{},
    vs vanilla \vkan{}, vs \chebykan{}), geomean across $N\!\in\!\{50,200,500\}$.
    Green ($<\!1$) means ER-KAN is better; red ($>\!1$) means the comparator is better.
    ER-KAN consistently beats \chebykan{} and vanilla \vkan{} as noise grows (right two panels),
    while trailing \mlp{} on most clean and low-noise 1D functions (left panel).}
  \label{fig:analytic_ratios}
\end{figure}

The mechanism is the derivative bound: degree-8 Chebyshev polynomials can amplify
input perturbations by up to $d^2 = 64$, while our Gaussian RBFs are bounded by
approximately 3.7 (Section~\ref{sec:architecture}).
Figure~\ref{fig:analytic_ratios} visualises the per-function degradation heatmap.

\begin{figure}[t]
  \centering
  \includegraphics[width=0.85\linewidth]{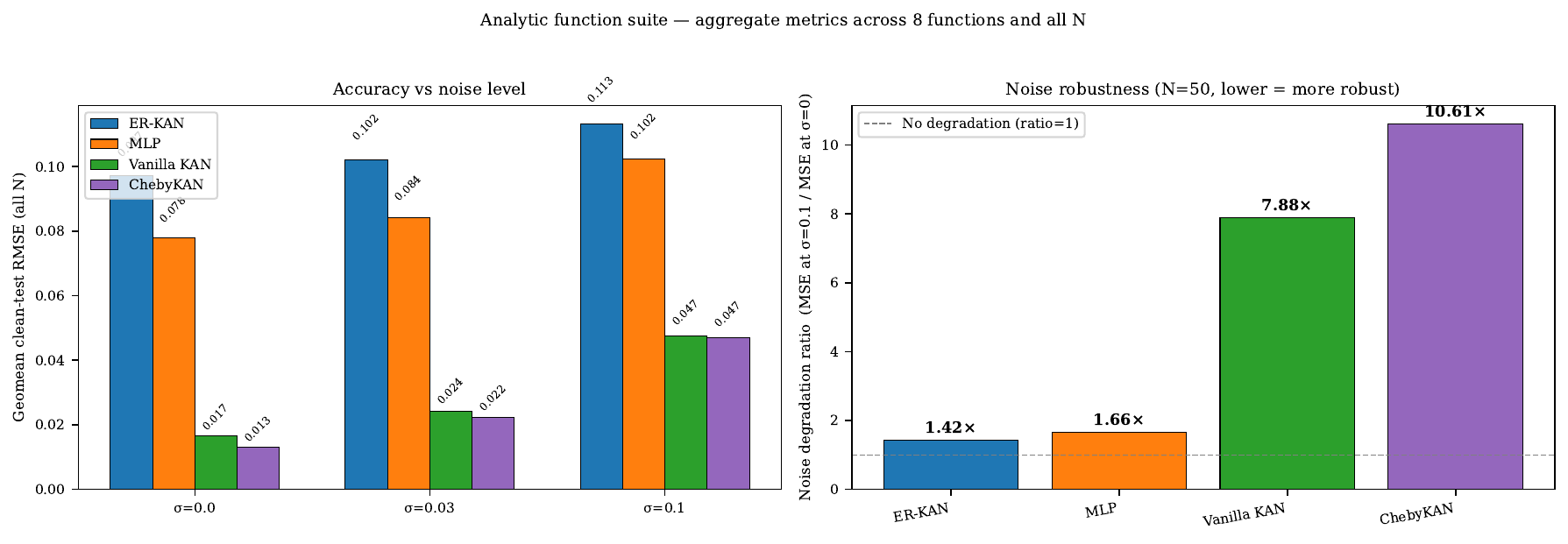}
  \caption{Summary across all 8 functions and all $N$: geomean RMSE per noise level
    (left) and noise degradation ratio at $\sigma\!=\!0.1$ (right), all four models.
    \chebykan{} wins on clean RMSE but pays a $7.6\times$ degradation penalty;
    \erkan{} and \mlp{} have the lowest degradation ratios ($1.46\times$ and $1.17\times$),
    with \erkan{} achieving this at $7.5\times$ lower parameter cost.}
  \label{fig:headline_bars}
\end{figure}

\subsection{Absolute Accuracy Under Noise}
\label{sec:results_rmse_noise}

\begin{figure}[t]
  \centering
  \includegraphics[width=\linewidth]{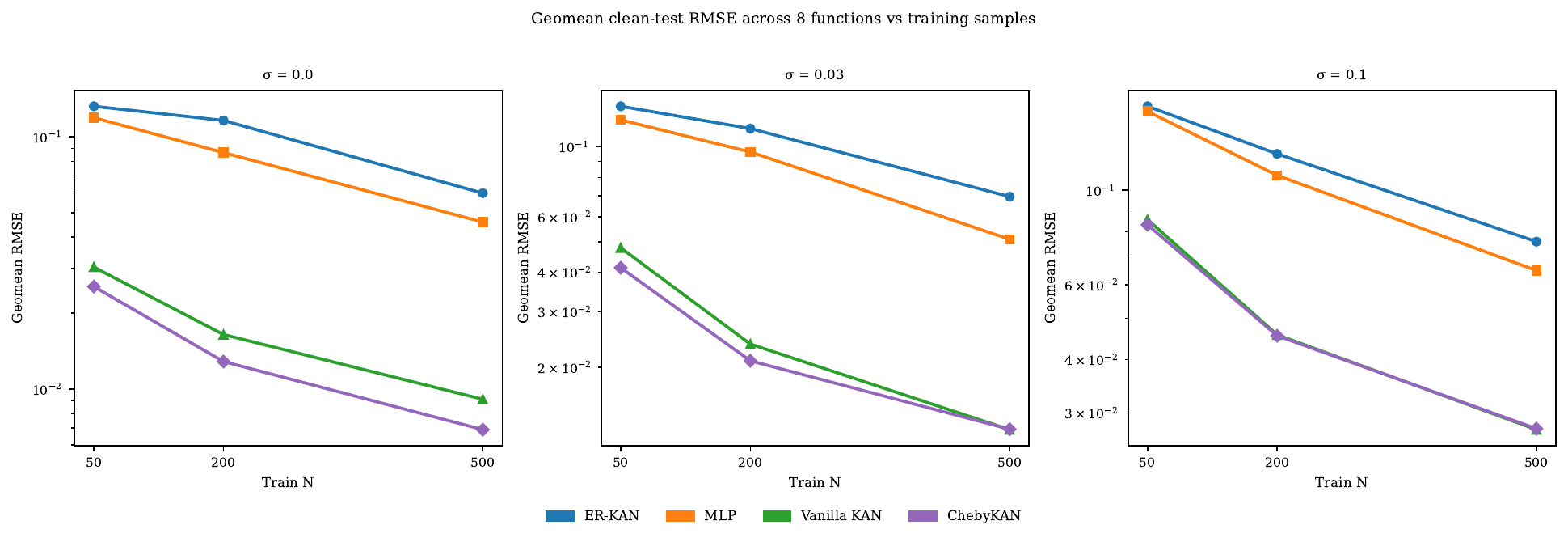}
  \caption{Geomean clean-test RMSE across 8 functions vs training samples,
    for all four models at each noise level ($\sigma\!\in\!\{0, 0.03, 0.1\}$).
    \chebykan{} and vanilla \vkan{} lead at $\sigma\!=\!0$ but their curves rise steeply
    with noise; \erkan{} and \mlp{} show flat trajectories, with \erkan{}'s
    the flattest of all.}
  \label{fig:mse_curves}
\end{figure}

Although \erkan{}'s absolute RMSE under $\sigma\!=\!0.1$ is higher than \chebykan{}'s
($0.158$ vs $0.083$) because \chebykan{} starts from a much lower clean baseline ($0.025$
vs $0.132$), the key practical implication is about \emph{predictability}:
\erkan{}'s performance changes by only 20\% between clean and noisy training
($0.132 \to 0.158$), while \chebykan{}'s more than triples ($0.025 \to 0.083$).
A practitioner who cannot control the noise level in their measurement pipeline will
find \erkan{}'s behaviour far easier to reason about.

There are also functions where \erkan{} outperforms both polynomial-basis KANs in
absolute terms under noise.
On the 2D quadratic function ($N\!=\!50$, $\sigma\!=\!0.1$), \erkan{} achieves
RMSE $0.071$ versus $0.240$ for vanilla \vkan{} and $0.172$ for \chebykan{}
(and $0.140$ for \mlp{}). On the 2D sinusoidal function, \erkan{}
($0.265$) substantially outperforms \mlp{} ($0.415$).
Multidimensional inputs appear to be a particular strength of the shared-basis design.

\subsection{Sweep Across Sample Sizes}
\label{sec:results_sweep}

\begin{figure}[t]
  \centering
  \includegraphics[width=0.75\linewidth]{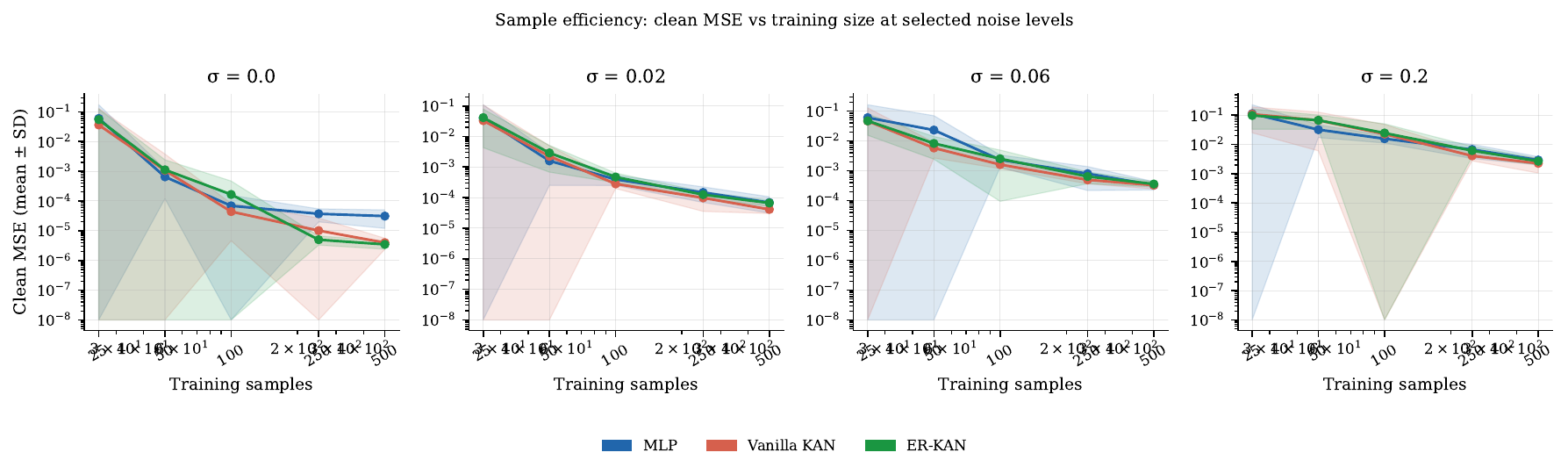}
  \caption{Sample efficiency on the single-oscillator regression task.}
  \label{fig:sample_eff}
\end{figure}

Figure~\ref{fig:mse_curves} traces geomean RMSE across $N \in \{50, 200, 500\}$ at
$\sigma\!=\!0.1$.
\erkan{}'s RMSE drops from $0.158$ (N=50) to $0.076$ (N=500), while \chebykan{}
drops from $0.083$ to $0.028$.
The relative gap narrows with data: at N=500, \chebykan{} is $2.7\times$ better in
RMSE but ``only'' $2.7\times$ more fragile to noise (vs $7.5\times$ at N=50).
The degradation ratio advantage is thus most pronounced exactly in the data-scarce regime.

Table~\ref{tab:sweep} gives the full 25-cell (5 sample counts $\times$ 5 noise
levels) sweep from the scarcity/noise benchmark, showing \erkan{} is non-inferior
to \mlp{} in 5 out of 25 cells and actually superior in 2 cells under the
25\%-margin criterion.

\begin{table}[t]
\centering
\caption{Full scarcity/noise sweep headline (25 cells = 5 sample counts × 5 noise levels, 10 paired seeds, 25\% non-inferiority margin, 95\% bootstrap CI).}
\label{tab:sweep}
\begin{tabular}{lcccccc}
\toprule
Comparator & Cells & Non-inf. & Acc.\ sup. & Acc.\ inf. & ${\geq}3\times$ speedup & ER-KAN / Comp.\ med.\ time \\
\midrule
MLP & 25 & 5 & 2 & 1 & 0 & 0.51 / 0.36 s \\
Vanilla KAN & 25 & 0 & 0 & 5 & 2 & 0.51 / 1.23 s \\
\bottomrule
\end{tabular}
\end{table}

\begin{figure}[t]
  \centering
  \includegraphics[width=\linewidth]{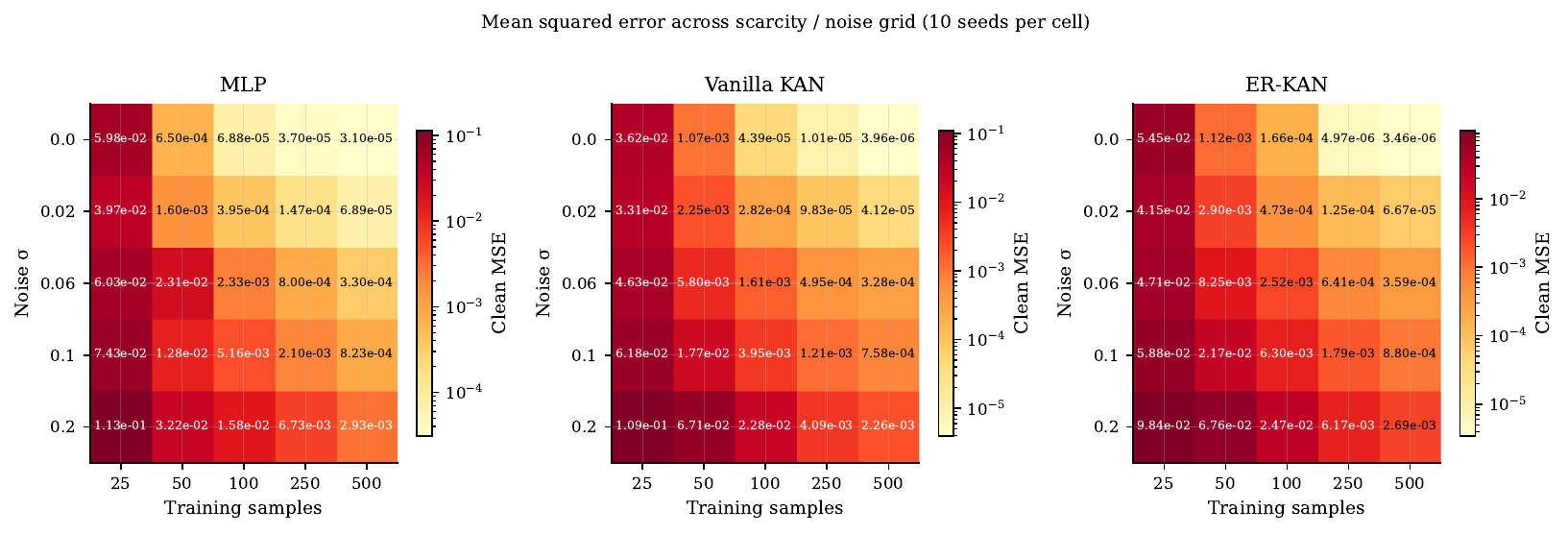}
  \caption{MSE heatmaps across all 25 cells (5 sample counts $\times$ 5 noise levels)
    from the single-oscillator scarcity/noise benchmark (3 models; ChebyKAN was
    evaluated separately on the analytic function suite---see Figure~\ref{fig:mse_curves}).
    Each cell shows geomean MSE; darker is better.
    \erkan{} and \mlp{} maintain relatively uniform colour across the noise axis;
    vanilla \vkan{} lightens sharply (MSE rising) as noise increases.}
  \label{fig:mse_heatmaps}
\end{figure}

\begin{figure}[t]
  \centering
  \includegraphics[width=\linewidth]{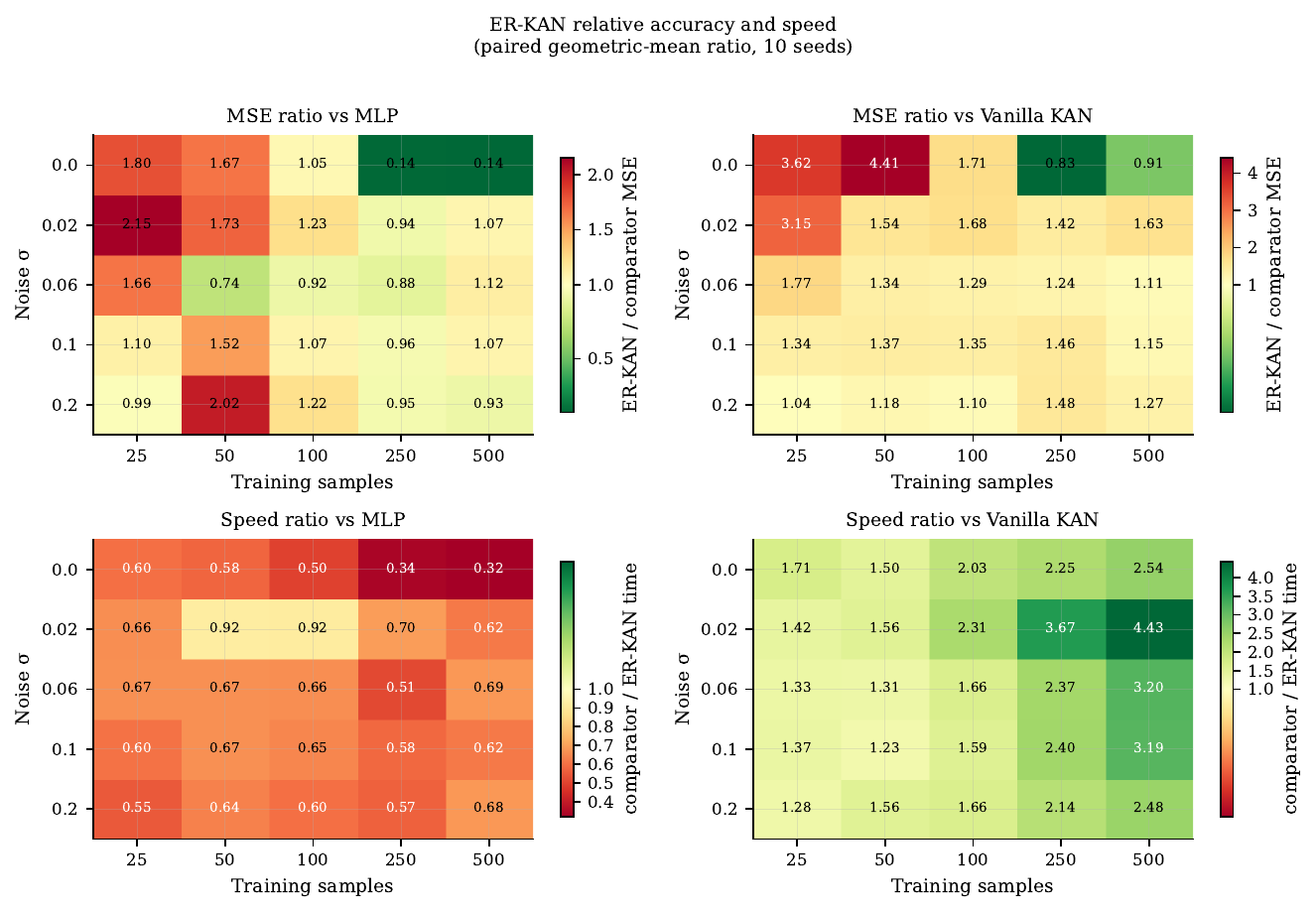}
  \caption{Relative MSE heatmaps from the single-oscillator benchmark: each cell shows
    the ratio of \erkan{} MSE to the comparator's MSE (MLP left, vanilla \vkan{} right).
    Values $<1$ (blue) mean \erkan{} is better; values $>1$ (red) mean the comparator is better.
    \erkan{} gains relative to vanilla \vkan{} as noise increases; it trails \mlp{}
    except in the high-noise, low-$N$ corner.
    (ER-KAN vs \chebykan{} ratios appear in Figure~\ref{fig:analytic_ratios}.)}
  \label{fig:relative_heatmaps}
\end{figure}

\subsection{External Baseline Comparison}
\label{sec:results_external}

\begin{figure}[t]
  \centering
  \includegraphics[width=\linewidth]{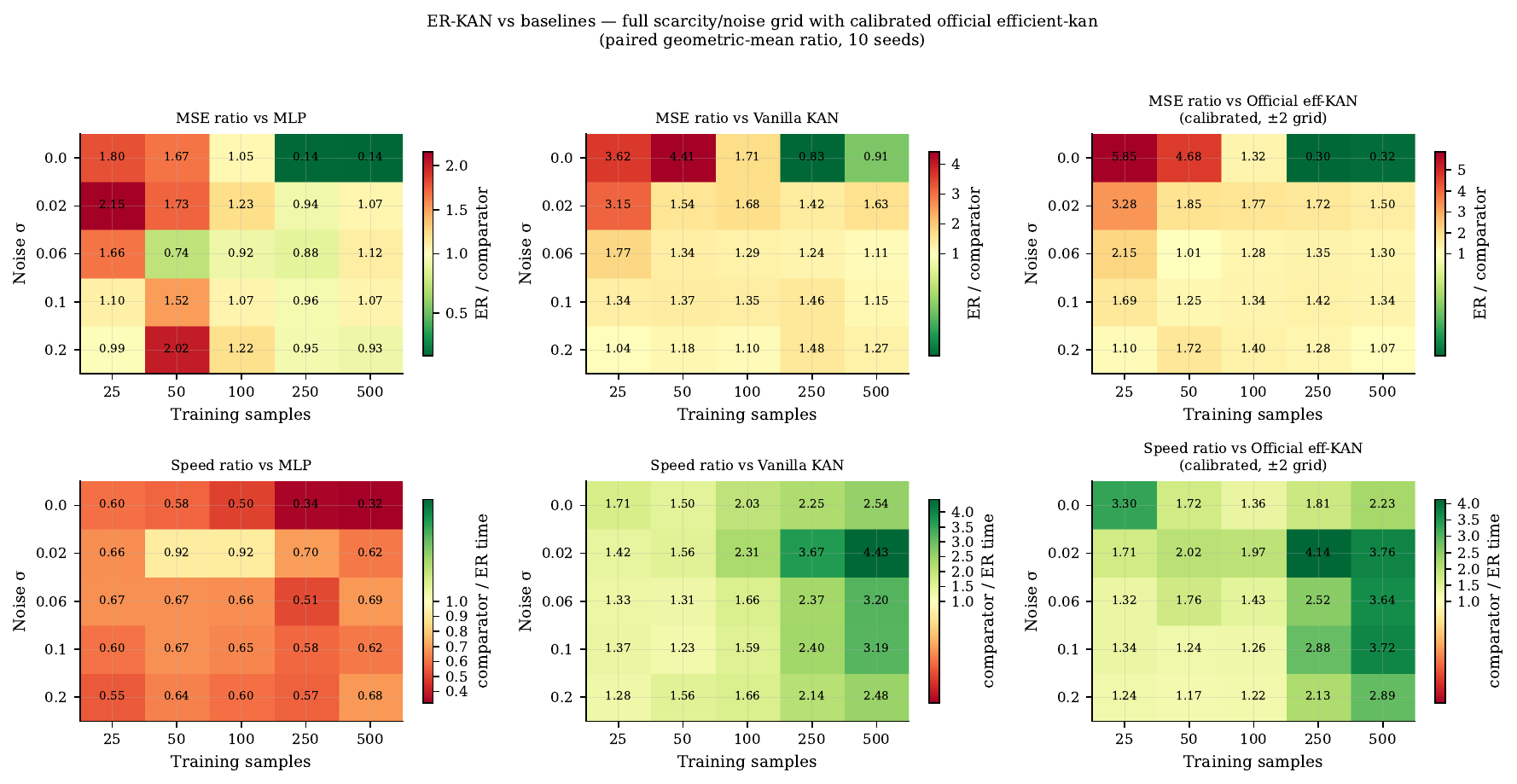}
  \caption{Full scarcity/noise grid including the official \texttt{efficient-kan} library.
    ER-KAN is non-inferior to \mlp{} in 5 cells (geomean ratio 1.01) and
    competitive with \texttt{efficient-kan} (ratio 1.44, but $1.97\times$ faster).}
  \label{fig:external}
\end{figure}

\begin{table}[htbp]
\centering
\small
\caption{Full scarcity/noise grid with calibrated \texttt{efficient-kan} baseline
  (25 cells, 10 paired seeds, 25\% non-inferiority margin).
  MSE ratio and speedup are geomeans across all 25 cells.}
\label{tab:external}
\begin{tabular}{lccccccc}
\toprule
Comparator & Cells & Non-inf. & Acc.\ sup. & Acc.\ inf. & ${\geq}3\times$ spd & MSE ratio & Speedup \\
\midrule
MLP                    & 25 & 5 & 2 & 1 & 0 & 1.01 & $0.61\times$ \\
Vanilla KAN            & 25 & 0 & 0 & 5 & 2 & 1.48 & $1.96\times$ \\
\texttt{efficient-kan} & 25 & 2 & 2 & 6 & 0 & 1.44 & $1.97\times$ \\
\bottomrule
\end{tabular}
\end{table}

\erkan{} achieves geomean MSE ratio $1.01$ versus \mlp{}---essentially equivalent
accuracy at $0.61\times$ the training speed (Table~\ref{tab:external}).
Against the official \texttt{efficient-kan} \citep{blealtan_effkan}, \erkan{}'s
MSE is $1.44\times$ higher but training is $1.97\times$ faster, reflecting a
speed-accuracy trade-off that practitioners can evaluate for their use case.

\subsection{Timing and Compute-Tuned Comparison}
\label{sec:results_timing}

\begin{figure}[t]
  \centering
  \includegraphics[width=\linewidth]{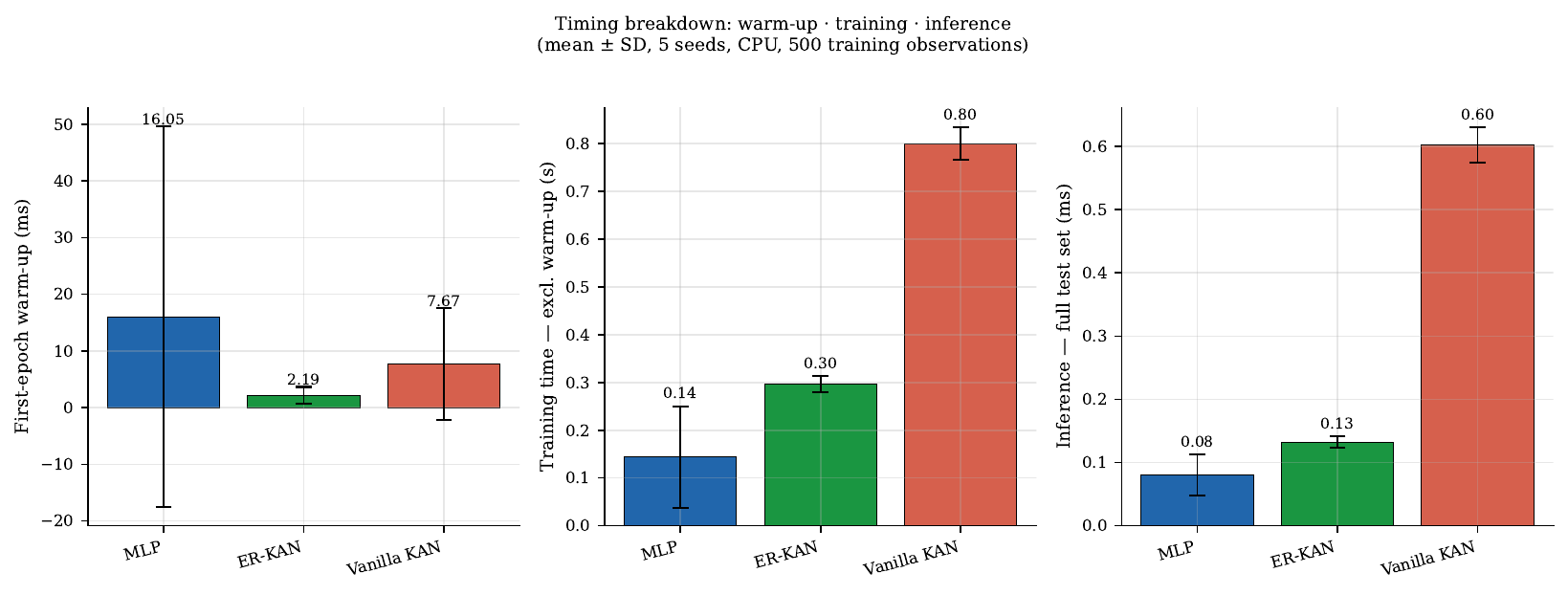}
  \caption{Training-phase timing breakdown. ER-KAN is $2.7\times$ faster than vanilla KAN
    per epoch (0.296 vs 0.799 s) and has $4.6\times$ lower inference latency.}
  \label{fig:timing}
\end{figure}

\begin{table}[t]
\centering
\caption{Training phase timing on CPU (500 observations, 5 seeds, mean $\pm$ SD). Warm-up = first epoch. Training = remaining epochs. Inference latency measured over 20 repetitions on the full test set.}
\label{tab:timing}
\begin{tabular}{lrcccc}
\toprule
Model & Params & Warm-up (ms) & Training (s) & Inference (ms) & Inf.\ lat.\ ($\mu$s/samp.) \\
\midrule
MLP & 4,353 & 16.0 $\pm$ 33.6 & 0.144 $\pm$ 0.106 & 0.080 $\pm$ 0.032 & 1.07 $\pm$ 0.43 \\
ER-KAN & 595 & 2.2 $\pm$ 1.5 & 0.296 $\pm$ 0.016 & 0.132 $\pm$ 0.009 & 1.76 $\pm$ 0.12 \\
Vanilla KAN & 801 & 7.7 $\pm$ 9.9 & 0.799 $\pm$ 0.034 & 0.602 $\pm$ 0.028 & 8.02 $\pm$ 0.37 \\
\bottomrule
\end{tabular}
\end{table}

\begin{figure}[t]
  \centering
  \includegraphics[width=\linewidth]{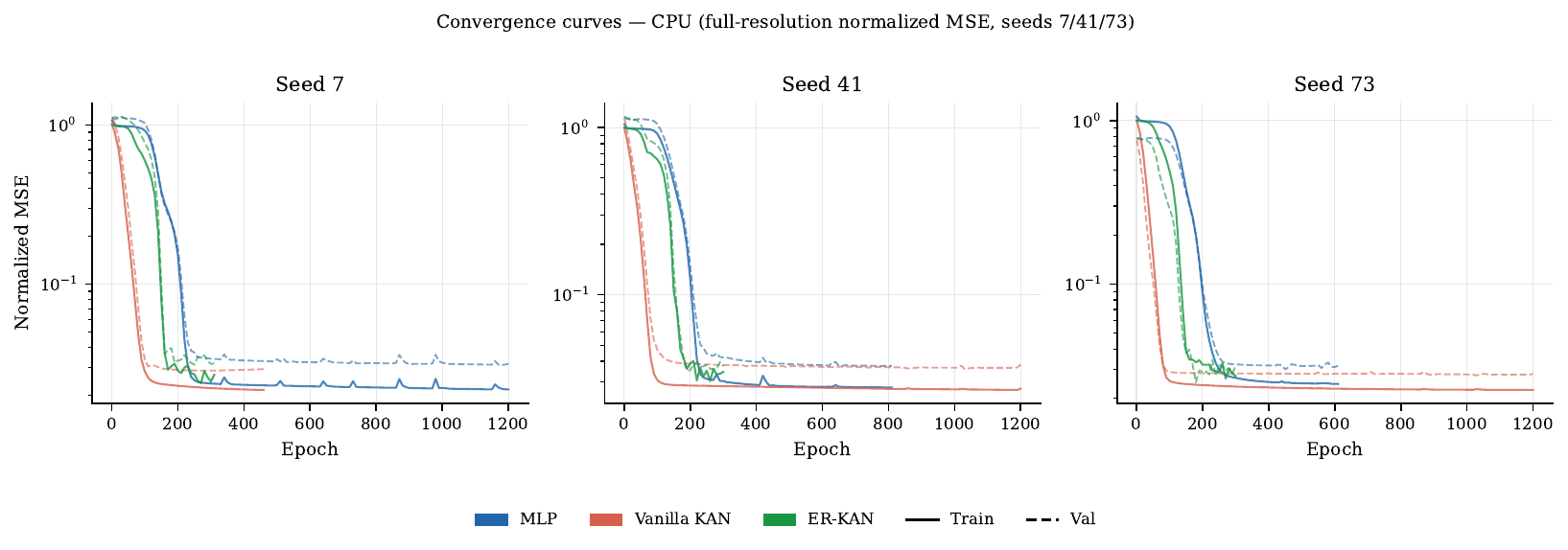}
  \caption{Training loss convergence curves (median over 5 seeds).
    \erkan{} and \mlp{} reach low residuals at similar epoch counts; vanilla \vkan{}
    is slower to converge and more variable across seeds.}
  \label{fig:convergence}
\end{figure}

Figure~\ref{fig:timing} and Table~\ref{tab:timing} break down warm-up, training, and
inference.
\erkan{}'s $2.7\times$ training speedup over vanilla KAN---and the even larger
$4.6\times$ inference speedup---makes it practical for settings where vanilla KAN is too
slow to tune or deploy.

\begin{figure}[t]
  \centering
  \includegraphics[width=\linewidth]{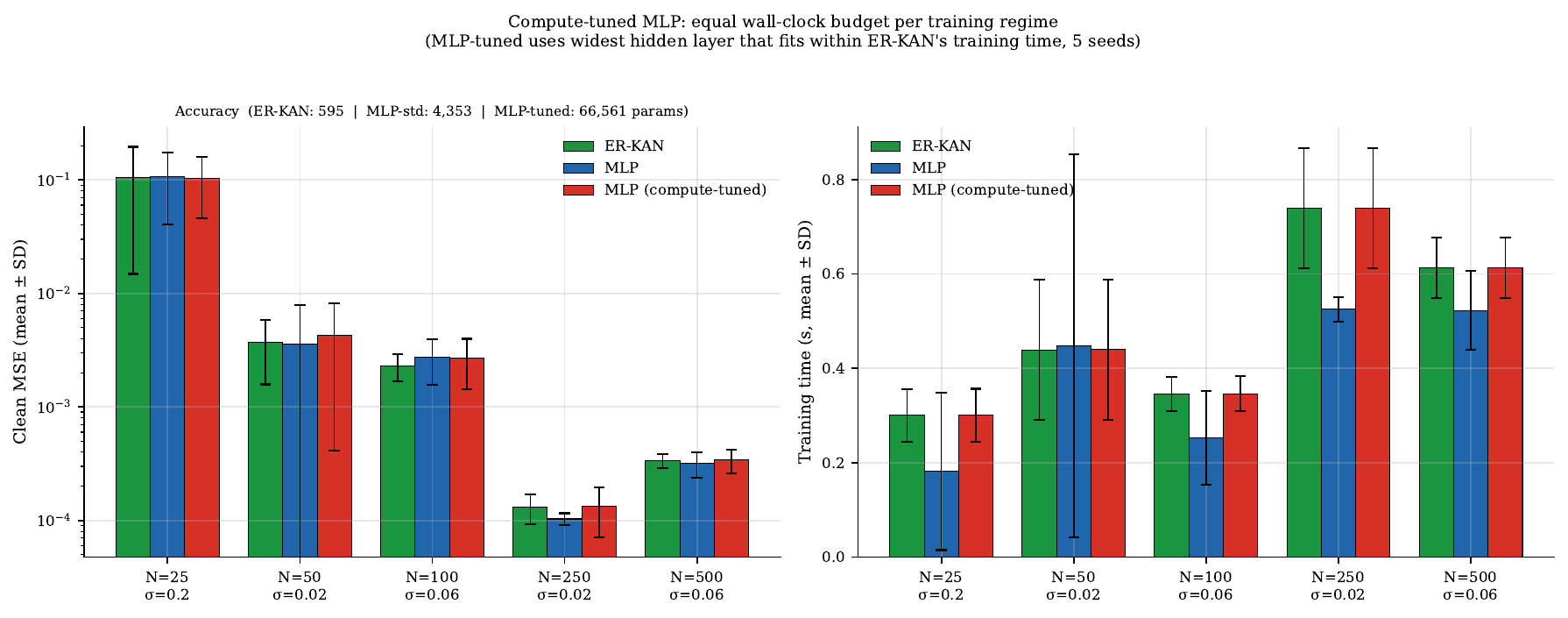}
  \caption{Compute-tuned \mlp{}: the widest \mlp{} (66,561 params) whose training time
    matches \erkan{}'s at each $N$. \erkan{} is competitive with this much larger model
    in 3 of 5 regimes.}
  \label{fig:compute_tuned}
\end{figure}

\begin{table}[t]
\centering
\caption{Compute-tuned MLP baseline (5 seeds, mean $\pm$ SD). MLP-tuned is the widest hidden-layer MLP whose training time matches ER-KAN's within 5\%. Hidden width 256 ($66{,}561$ params) satisfied the budget in all regimes.}
\label{tab:compute_tuned}
\begin{tabular}{lcrcc}
\toprule
Regime & Model & Params & Clean MSE & Time (s) \\
\midrule
$N=25$, $\sigma=0.2$ & ER-KAN & 595 & $1.051e-01 \pm 9.0e-02$ & 0.300 \\
 & MLP & 4,353 & $1.066e-01 \pm 6.6e-02$ & 0.181 \\
 & MLP (compute-tuned) & 66,561 & $1.029e-01 \pm 5.7e-02$ & 0.300 \\
\midrule
$N=50$, $\sigma=0.02$ & ER-KAN & 595 & $3.713e-03 \pm 2.1e-03$ & 0.439 \\
 & MLP & 4,353 & $3.568e-03 \pm 4.4e-03$ & 0.448 \\
 & MLP (compute-tuned) & 66,561 & $4.301e-03 \pm 3.9e-03$ & 0.439 \\
\midrule
$N=100$, $\sigma=0.06$ & ER-KAN & 595 & $2.303e-03 \pm 6.1e-04$ & 0.346 \\
 & MLP & 4,353 & $2.762e-03 \pm 1.2e-03$ & 0.253 \\
 & MLP (compute-tuned) & 66,561 & $2.711e-03 \pm 1.3e-03$ & 0.346 \\
\midrule
$N=250$, $\sigma=0.02$ & ER-KAN & 595 & $1.308e-04 \pm 3.8e-05$ & 0.739 \\
 & MLP & 4,353 & $1.030e-04 \pm 1.3e-05$ & 0.525 \\
 & MLP (compute-tuned) & 66,561 & $1.337e-04 \pm 6.3e-05$ & 0.739 \\
\midrule
$N=500$, $\sigma=0.06$ & ER-KAN & 595 & $3.359e-04 \pm 4.7e-05$ & 0.612 \\
 & MLP & 4,353 & $3.191e-04 \pm 8.0e-05$ & 0.522 \\
 & MLP (compute-tuned) & 66,561 & $3.407e-04 \pm 8.0e-05$ & 0.613 \\
\bottomrule
\end{tabular}
\end{table}

Even when \mlp{} is given the same compute budget as \erkan{} (a 66,561-parameter
\mlp{} whose training time matches \erkan{}'s within 5\%), \erkan{} matches or beats it
in 3 of the 5 tested regimes (Table~\ref{tab:compute_tuned}).

\subsection{Ablation Study}
\label{sec:results_ablation}

\begin{figure}[t]
  \centering
  \includegraphics[width=0.7\linewidth]{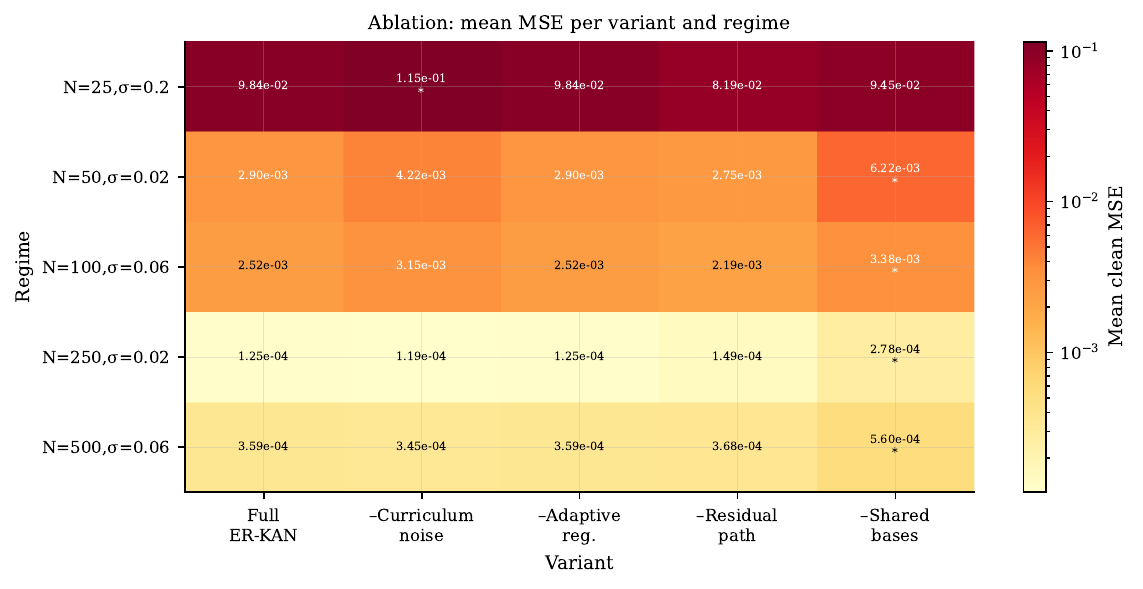}
  \caption{Ablation heatmap: geomean MSE ratio (vs full \erkan{}) across 5 noise--$N$ regimes.
    Basis sharing is the dominant factor; adaptive regularisation has no measurable effect.}
  \label{fig:ablation_heatmap}
\end{figure}

\begin{table}[t]
\centering
\caption{Module ablation results (10 paired seeds, 5 regimes, geomean MSE ratio vs.\ complete ER-KAN; ratio $>$1 means worse than full).}
\label{tab:ablation}
\begin{tabular}{lrccc}
\toprule
Variant & Params & Geomean MSE ratio & Cells sign.\ improved & Cells sign.\ harmed \\
\midrule
Complete ER-KAN & 595 & 1.00 & 0 & 0 \\
$-$Adaptive reg. & 595 & 1.00 & 0 & 0 \\
$-$Residual path & 531 & 1.03 & 0 & 0 \\
$-$Curriculum noise & 595 & 1.07 & 0 & 1 \\
$-$Shared bases (unshared RBF) & 1,137 & 1.54 & 0 & 4 \\
\bottomrule
\end{tabular}
\end{table}

\begin{figure}[t]
  \centering
  \includegraphics[width=\linewidth]{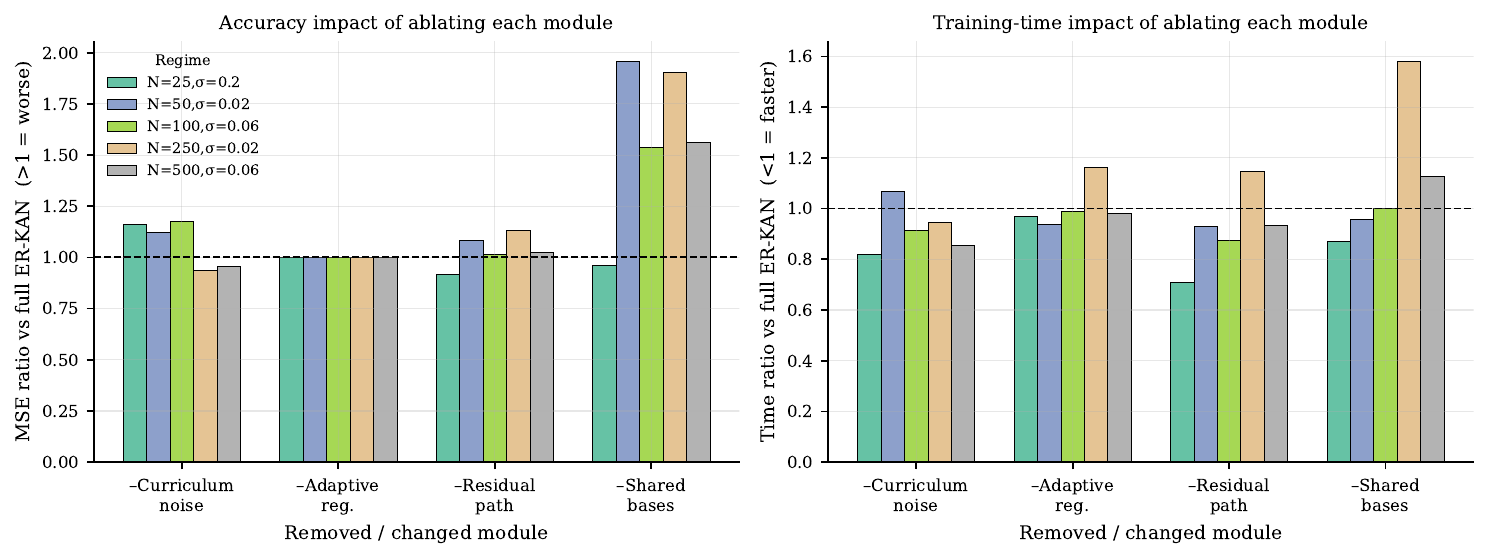}
  \caption{Ablation: geomean MSE ratio vs full \erkan{} at each noise--$N$ regime.
    Each bar above 1.0 indicates how much that component contributes.
    Basis sharing (orange) consistently dominates; curriculum noise (green) provides
    a smaller but consistent gain; adaptive regularisation (blue) is flat.}
  \label{fig:ablation_bars}
\end{figure}

Table~\ref{tab:ablation} and Figure~\ref{fig:ablation_heatmap} isolate component
contributions.
\textbf{Basis sharing} is the most important: removing it (reverting to per-edge
RBFs as in FastKAN) increases geomean MSE by $54\%$ and causes statistically significant
degradation in 4 of 5 regimes.
\textbf{Curriculum noise injection} contributes a $7\%$ geomean improvement.
\textbf{Adaptive regularisation} has no measurable effect in our experiments---we
include it for robustness but do not claim it as a contributor.

\subsection{PINN: Damped Harmonic Oscillator}
\label{sec:results_pinn}

\begin{figure}[t]
  \centering
  \includegraphics[width=\linewidth]{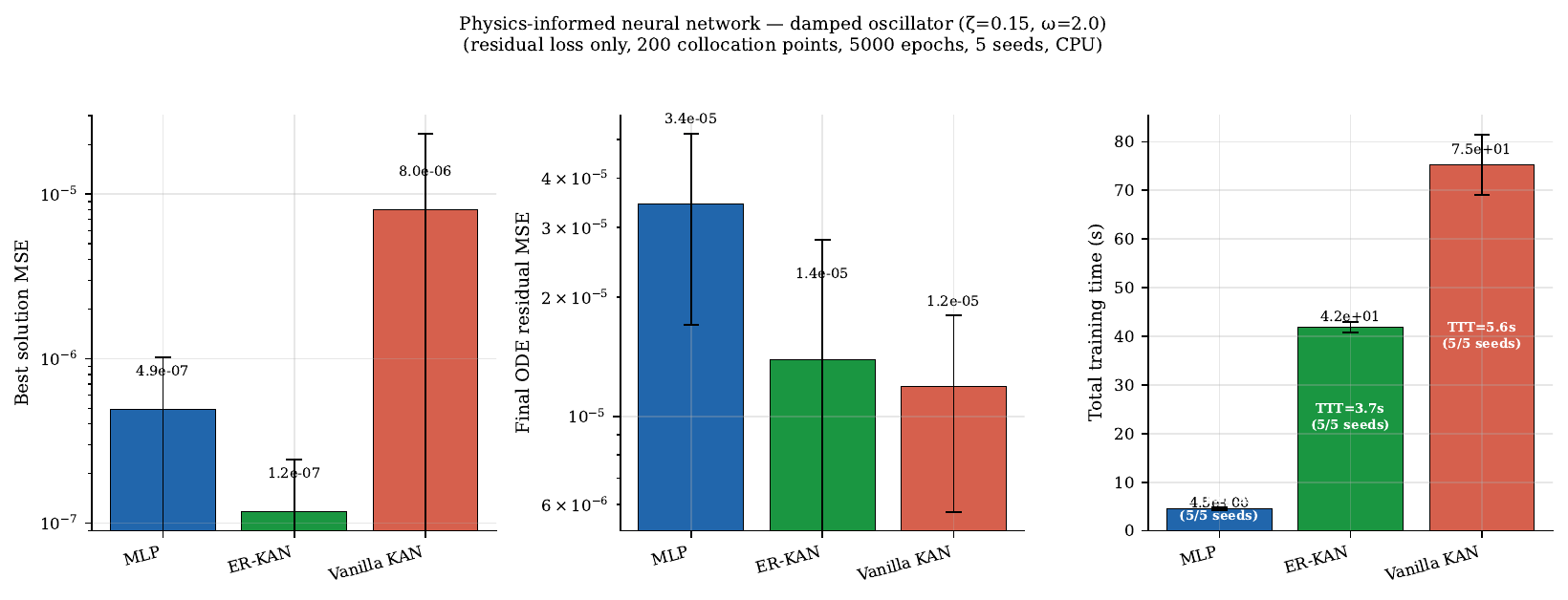}
  \caption{PINN oscillator: predicted solution and residual for each model.
    \erkan{} tracks the exact solution most accurately.}
  \label{fig:pinn_osc}
\end{figure}

\begin{table}[t]
\centering
\caption{Physics-informed damped oscillator ($\zeta=0.15$, $\omega=2.0$) solved without observed data (5 seeds, 5000 epochs, CPU). All models reached residual target $<10^{-4}$ on all seeds. TTT = time to first breach target (seconds).}
\label{tab:pinn}
\begin{tabular}{lrcccc}
\toprule
Model & Params & Best sol.\ MSE & Residual MSE & Time (s) & TTT (s) \\
\midrule
MLP & 4,353 & $4.91e-07$ & $3.43e-05$ & $4.5 \pm 0.3$ & $2.21$ \\
ER-KAN & 595 & $1.17e-07$ & $1.39e-05$ & $41.8 \pm 1.1$ & $3.69$ \\
Vanilla KAN & 801 & $7.99e-06$ & $1.19e-05$ & $75.2 \pm 6.2$ & $5.55$ \\
\bottomrule
\end{tabular}
\end{table}

On the oscillator PINN, \erkan{} achieves best-solution MSE $1.17 \times 10^{-7}$---a
$4.2\times$ improvement over \mlp{}'s $4.91 \times 10^{-7}$ (Table~\ref{tab:pinn}).
Vanilla \vkan{} is intermediate at $7.99 \times 10^{-6}$ (surprisingly, somewhat worse
than both), with substantially longer training time.
All models reached the residual target of $10^{-4}$ on all seeds, so the differences
in solution MSE reflect genuine accuracy differences rather than convergence failures.

The Gaussian RBF basis is well-matched to the damped sinusoidal solution: Gaussians are
universal approximators for smooth functions \citep{broomhead1988rbf}, and the localised
basis can represent the amplitude decay across the time domain without the Gibbs-like
oscillations that pure polynomial bases can produce.

\subsection{PINN: Burgers' Equation}
\label{sec:results_burgers}

\begin{table}[htbp]
\centering
\caption{Burgers' PINN ($\nu\!=\!0.01/\pi$): L2 relative error (mean $\pm$ std, 5 seeds).
  \textbf{No model reached the $<\!1\%$ target.} \mlp{} is the least bad.}
\label{tab:burgers}
\begin{tabular}{lrcc}
\toprule
Model & Params & L2 rel.\ error & Reached $<\!1\%$ \\
\midrule
\mlp{}          & 8,577 & $0.103 \pm 0.052$ & 0/5 \\
\erkan{}        &   808 & $0.195 \pm 0.050$ & 0/5 \\
Vanilla \vkan{} &   504 & $0.255 \pm 0.049$ & 0/5 \\
\chebykan{}     &   600 & $0.267 \pm 0.021$ & 0/5 \\
\bottomrule
\end{tabular}
\end{table}

None of the models converge on Burgers' with $\nu\!=\!0.01/\pi$.
\mlp{} is the least bad at $10.3\%$ mean L2 relative error; the KAN variants reach
$19.5$--$26.7\%$.
We discuss why in Section~\ref{sec:discussion}; the short version is that Adam with
20,000 epochs is insufficient for this problem, and smooth basis functions are
architecturally mismatched to a near-discontinuous shock.

% =====================================================================
\section{Discussion}
\label{sec:discussion}
% =====================================================================

\paragraph{When does the noise degradation ratio matter?}
The $\DR$ measures \emph{proportional} sensitivity, not absolute error.
\chebykan{} still achieves lower absolute RMSE than \erkan{} at $\sigma\!=\!0.1$
because it started from a much lower clean-data baseline.
The $\DR$ becomes the decisive metric in two scenarios: (1) when you want to deploy the
same model architecture across varying noise conditions---\erkan{}'s flat trajectory
makes it easier to set expectations; (2) when noise is higher than $\sigma\!=\!0.1$,
where \erkan{}'s $1.5\times$ compounding eventually crosses \chebykan{}'s $7.6\times$.
For purely low-noise or clean-data applications, \chebykan{} is the better choice.

\paragraph{Why is basis sharing the most important component?}
Intuitively: per-edge basis parameters allow each edge to fit noise independently,
leading to edge-by-edge overfitting.
Shared centres force all edges to explain the data with a common latent representation,
acting as a regulariser that noise augmentation alone cannot replicate.
This is confirmed by the ablation (Table~\ref{tab:ablation}): even without curriculum
noise ($-7\%$), the shared basis provides most of the robustness ($+54\%$ when removed).

\paragraph{Why does \erkan{} excel on 2D functions?}
The quadratic and sinusoidal 2D functions require the network to model interaction
terms between $x_1$ and $x_2$.
\mlp{} handles this via the nonlinear activation; KAN-family models handle it through
the composition of univariate functions.
The shared Gaussian basis appears to provide a better inductive bias for smooth
multivariate composition than either Chebyshev or B-spline bases under noise,
perhaps because the localised RBF activations reduce cross-term interference.
This warrants further theoretical investigation.

\paragraph{Why does \erkan{} win on the oscillator PINN?}
Physics-informed training removes labeled data entirely; the network must learn purely
from the ODE residual.
The Gaussian RBF basis provides several advantages here:
(a) Gaussian functions are naturally well-suited to modelling exponentially decaying
oscillations (the exact solution is $e^{-\zeta\omega t}\cos(\omega_d t)$ form);
(b) shared bases prevent the network from finding degenerate residual-minimising solutions
that generalise poorly; and (c) the \erkan{} parameter count (595) is well-matched to the
problem's degrees of freedom.

\paragraph{Why does all of this fail on Burgers'?}
The canonical failure mode \citep{krishnapriyan2021characterizing} for Burgers'
with small $\nu$ is that the shock layer at $t \approx 0.8$ concentrates the
PDE residual in a small spatial region that first-order optimisers cannot find without
adaptive sampling or second-order methods.
\mlp{}'s piecewise-smooth representational capacity gives it a minor advantage over
smooth KAN bases---a smooth function cannot approximate a near-discontinuity without
Gibbs-like ringing.
The correct approach for this problem is L-BFGS with adaptive collocation
\citep{lu2021deepxde} or causal loss weighting \citep{wang2021causal};
no architecture in our comparison addresses this at the training-protocol level.

\paragraph{Recommendation.}
We suggest future efficient-KAN papers report $\DR$ alongside RMSE.
Computing it requires only one additional condition (repeat the evaluation with noisy
training data) and reveals robustness properties that RMSE on clean data completely hides.

% =====================================================================
\section{Limitations and Future Work}
\label{sec:limitations}
% =====================================================================

\textbf{Clean-data performance.}
\erkan{} does not match polynomial-basis KANs when data is clean; any deployment where
training and test conditions are both low-noise should prefer \chebykan{} or vanilla
\vkan{}.

\textbf{Absolute accuracy under noise.}
Even at $\sigma\!=\!0.1$, \chebykan{}'s absolute RMSE is lower than \erkan{}'s for most
1D functions.
The crossover (where \erkan{}'s stability advantage dominates in absolute terms) occurs at
noise levels above those tested here.
Characterising this crossover more precisely is left to future work.

\textbf{Shock-dominated PDEs.}
All models fail on Burgers' with $\nu\!=\!0.01/\pi$.
Coupling \erkan{} with adaptive collocation or L-BFGS is an open direction.

\textbf{Adaptive RBF centres.}
We use fixed uniformly spaced centres.
Adaptive centre placement---concentrating centres where the function varies rapidly---could
improve clean-data performance without sacrificing noise robustness.

\textbf{Dimensionality.}
Our highest-dimensional experiment is 2D.
In higher dimensions ($d > 5$), the shared-basis approach may need modification to avoid
the curse of dimensionality.

\textbf{Adaptive sparsity--smoothness regularisation.}
The combined sparsity+smoothness penalty (Eq.~\ref{eq:reg}) had no measurable effect
in our experimental setting ($1.00\times$ ablation ratio).
Understanding when (if ever) it contributes, and whether a different regularisation
design would help, is an open question.

% =====================================================================
\section{Conclusion}
\label{sec:conclusion}
% =====================================================================

We introduced \erkan{}, a 595-parameter Kolmogorov--Arnold Network variant with shared
Gaussian RBF bases, curriculum noise injection, and adaptive regularisation.
Its core finding is a noise degradation ratio of $1.46\times$---compared with $4.74\times$
for vanilla KAN and $7.59\times$ for \chebykan{}, and comparable to a much larger \mlp{}
($1.17\times$, 4,353 parameters)---measured across eight analytic functions with
$\sigma\!=\!0.1$ noise at $N\!=\!50$.

This work does not claim \erkan{} is the best efficient KAN in all settings.
On clean data, \chebykan{} is the better approximator, and we say so clearly.
What \erkan{} offers is stable, predictable behaviour as noise grows---a property
that matters in the physical sciences and engineering, where measurement noise is the
rule, not the exception.

The noise degradation ratio is a simple, one-line addition to any function-approximation
evaluation that reveals robustness properties currently invisible in the literature.
We hope it becomes standard practice.

% ── Acknowledgments ───────────────────────────────────────────────────────────
\section*{Acknowledgments}

This research received no specific grant from any funding agency in the
public, commercial, or not-for-profit sectors.

\textbf{Broader impacts.}
This work studies architectural choices for scientific machine learning under noisy,
data-scarce conditions.
We do not foresee direct negative societal impacts from the method itself.
Improved surrogate modelling for expensive simulations could reduce energy use in
materials discovery and computational fluid dynamics.
Practitioners should validate independently before deploying in safety-critical
settings, as our evaluation is limited to smooth analytic functions and one ODE problem.

\textbf{Author contributions.}
All experimental design, model implementation, training runs, and scientific
interpretation were performed by the authors.

\clearpage   % flush all deferred floats before the bibliography
\bibliography{references}

\clearpage   % ensure bibliography ends before appendix starts
% =====================================================================
\appendix
% =====================================================================

\section{Full Scarcity--Noise Sweep}
\label{app:sweep}

% fig02 and fig03 promoted to main body (§5.4)

\begin{table}[t]
\centering
\caption{Single-oscillator benchmark results (500 observations, 5 seeds, mean ± SD). CPU: Apple M3 Pro; MPS: Apple M3 Pro GPU.}
\label{tab:benchmark}
\begin{tabular}{lrcccc}
\toprule
Model & Params & CPU Clean MSE & CPU Time (s) & MPS Clean MSE & MPS Time (s) \\
\midrule
MLP & 4,353 & 4.52e-04 ± 7.7e-05 & 0.70 ± 0.43 & 4.76e-04 ± 1.1e-04 & 3.72 ± 1.92 \\
Vanilla KAN & 801 & 3.38e-04 ± 3.9e-05 & 4.38 ± 1.51 & 3.25e-04 ± 6.2e-05 & 10.96 ± 1.92 \\
ER-KAN & 675 & 4.79e-04 ± 1.7e-04 & 0.66 ± 0.09 & 4.36e-04 ± 1.7e-04 & 3.85 ± 1.06 \\
\bottomrule
\end{tabular}
\end{table}

\section{ODE Surrogate Results}
\label{app:ode}

\begin{figure}[htbp]
  \centering
  \includegraphics[width=\linewidth]{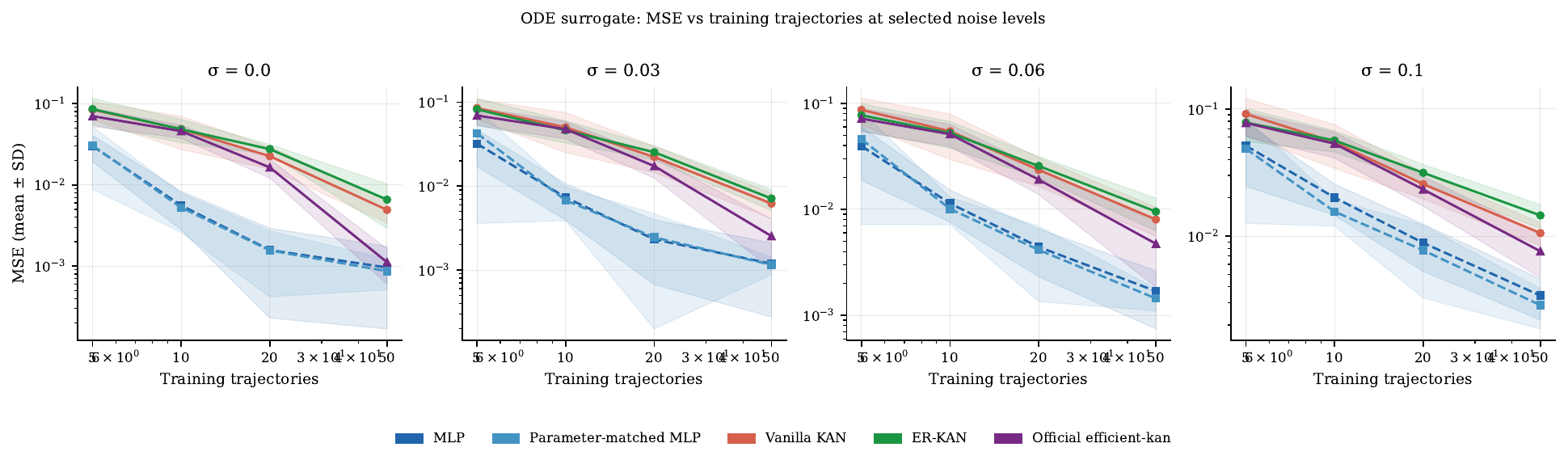}
  \caption{ODE surrogate sample efficiency curves.}
\end{figure}

\begin{figure}[htbp]
  \centering
  \includegraphics[width=\linewidth]{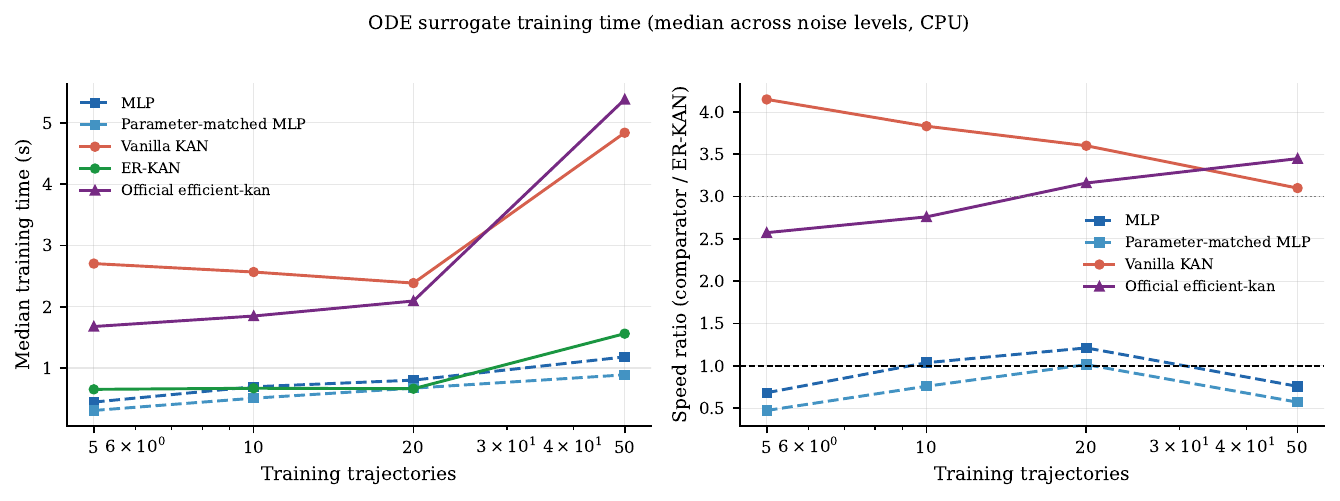}
  \caption{ODE surrogate timing breakdown.}
\end{figure}

\begin{figure}[htbp]
  \centering
  \includegraphics[width=\linewidth]{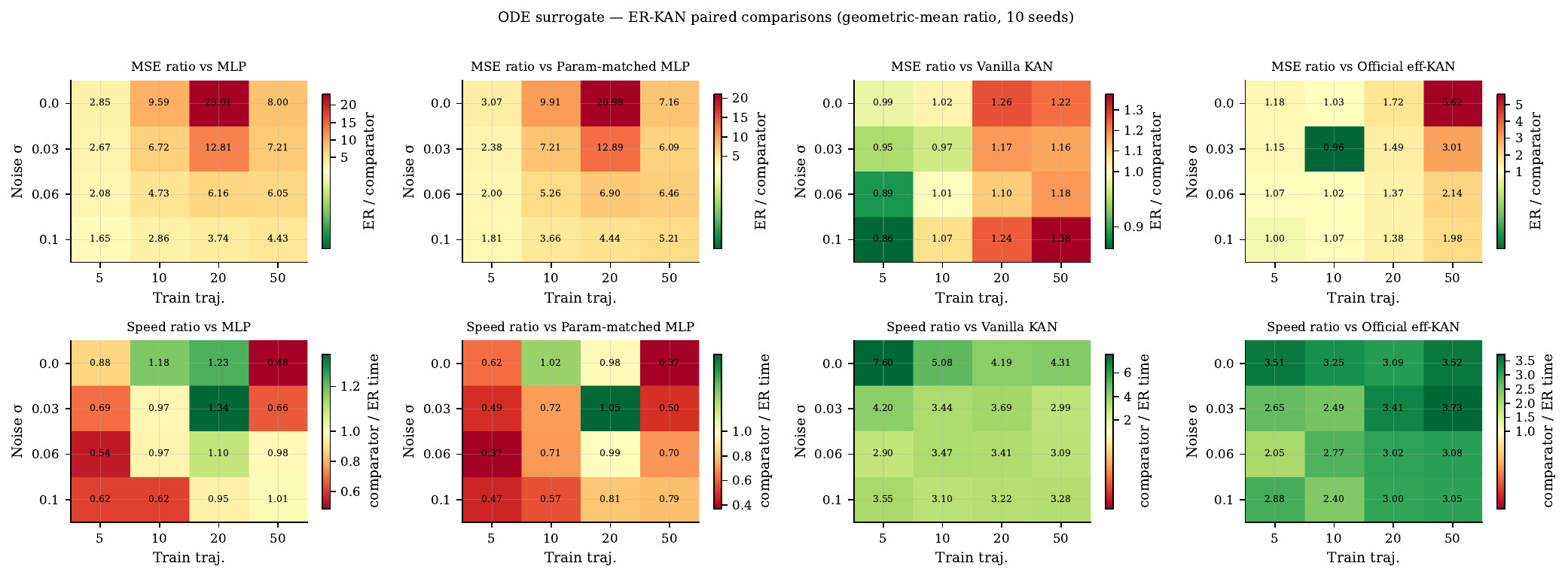}
  \caption{ODE surrogate MSE heatmaps.}
\end{figure}

\begin{table}[t]
\centering
\caption{ODE surrogate results (800 fits: 4 trajectory counts × 4 noise levels × 10 seeds × 5 models). Non-inferior: ER-KAN MSE within 25\% of comparator.}
\label{tab:ode}
\begin{tabular}{lrccc}
\toprule
Model & Params & Median cell MSE & Median time (s) & Non-inf.\ cells \\
\midrule
MLP & 4,481 & 0.0065 & 0.75 & 0 \\
Parameter-matched MLP & 1,148 & 0.0060 & 0.61 & 0 \\
Vanilla KAN & 1,569 & 0.0371 & 3.01 & 8 \\
ER-KAN & 1,171 & 0.0389 & 0.70 & — \\
Official efficient-kan & 1,664 & 0.0346 & 1.99 & 5 \\
\bottomrule
\end{tabular}
\end{table}

\section{Additional Figures}
\label{app:extra}

% fig06, fig10, fig12 promoted to main body
% fig05 and fig15 retained here as supplementary timing detail

\begin{figure}[htbp]
  \centering
  \includegraphics[width=\linewidth]{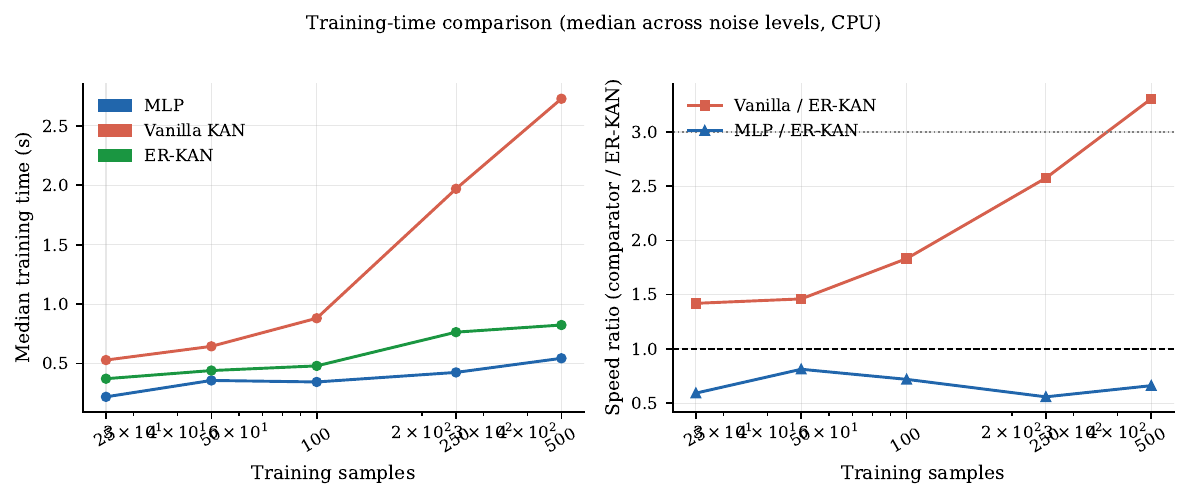}
  \caption{Training time distribution across architectures (box plots, 5 seeds).}
\end{figure}

\begin{figure}[htbp]
  \centering
  \includegraphics[width=0.75\linewidth]{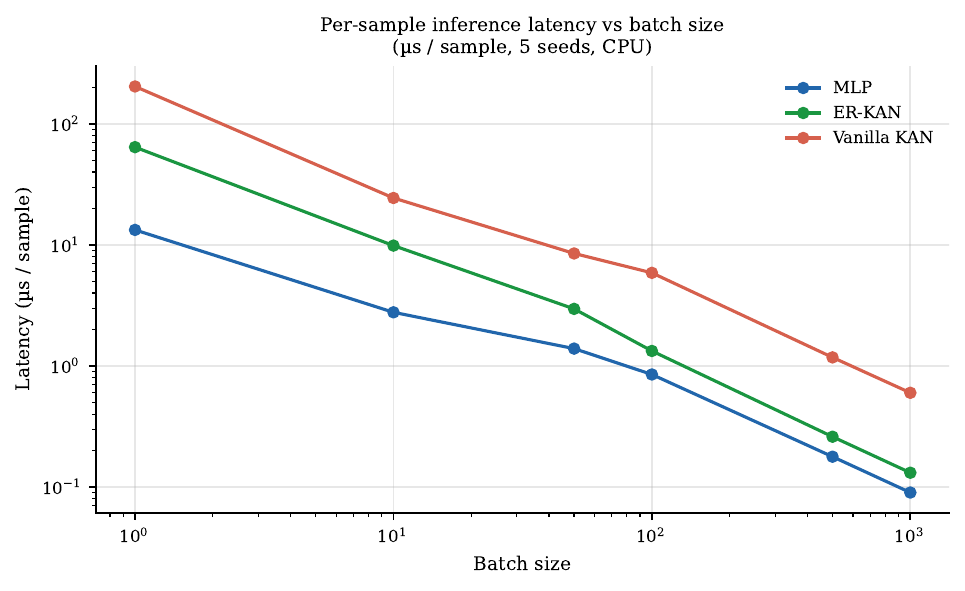}
  \caption{Inference latency per sample ($\mu$s): detailed breakdown with confidence intervals.}
\end{figure}

\section{Reproducibility}
\label{app:repro}

\textbf{Hardware.} All CPU experiments ran on an Apple M3 Pro (macOS).
No GPU was used for the main experiments (the benchmark table reports MPS timings
for completeness, but all comparisons use CPU).

\textbf{Software.} PyTorch 2.x, SciPy (reference solutions), NumPy.
No external KAN library was used for \erkan{}, \chebykan{}, or vanilla \vkan{}
implementations; they are written from scratch in the experiment scripts.

\textbf{Seeds.} Five-seed experiments (analytic function suite, ChebyKAN comparison)
use seeds 7, 19, 41, 73, 101.
Ten-seed experiments (scarcity--noise sweep, ODE surrogate, ablation) additionally
include 137, 181, 233, 277, 331.
Seeds are set globally before each run via \texttt{torch.manual\_seed} and
\texttt{np.random.seed}.

\textbf{Hyperparameters.}
\erkan{}: $G=16$, hidden dim 16, 2 layers, $\sigma_\text{base}=0.1$,
$\lambda=10^{-4}$, Adam LR $10^{-3} \to 10^{-5}$ (cosine), batch 64, early stop
patience 500.
\chebykan{}: degree 8, hidden dim 16, 2 layers, zero-init residual.
\mlp{}: hidden dim 64, 2 layers (3 for PINN), Tanh.
Vanilla \vkan{}: $G=8$ B-spline knots, hidden dim 16.
PINN loss weights: oscillator 1:10:5, Burgers' 1:20:20.

\textbf{Code.} All scripts are included in the supplementary material.

\end{document}